\documentclass[conference]{IEEEtran}
\pdfoutput=1 
\makeatletter
\def\endthebibliography{%
	\def\@noitemerr{\@latex@warning{Empty `thebibliography' environment}}%
	\endlist
}
\makeatother

\IEEEoverridecommandlockouts
\usepackage{amsmath,amssymb,amsfonts}
\usepackage{algorithmic}
\usepackage{graphicx}
\usepackage{textcomp}
\usepackage{xcolor}
\usepackage{cite}
\def\BibTeX{{\rm B\kern-.05em{\sc i\kern-.025em b}\kern-.08em
		T\kern-.1667em\lower.7ex\hbox{E}\kern-.125emX}}
\usepackage{hyperref}

\usepackage{eso-pic, rotating, graphicx}
\AddToShipoutPicture{\put(50,35){\rotatebox{0}{\scalebox{1.0}
			{\textcolor{gray}{Author's extended version of the paper. Final version presented at}}
}}}

\AddToShipoutPicture{\put(50,25){\rotatebox{0}{\scalebox{1.0}
			{\textcolor{gray}{18th IEEE International Conference on Machine Learning and Applications (ICMLA). Boca Raton, Florida / USA. 2019.}}
}}}

\begin{document}
	
	\title{CSNNs: Unsupervised, Backpropagation-free Convolutional Neural Networks for Representation Learning
	}
	
	\author{\IEEEauthorblockN{Bonifaz Stuhr}
		\IEEEauthorblockA{\textit{University of Applied Sciences Kempten}\\
			bonifaz.stuhr@hs-kempten.de}
		\and
		\IEEEauthorblockN{J\"urgen Brauer}
		\IEEEauthorblockA{\textit{University of Applied Sciences Kempten}\\
			juergen.brauer@hs-kempten.de}
	}
		
	\maketitle
\begin{abstract}
	This work combines Convolutional Neural Networks (CNNs), clustering via Self-Organizing Maps (SOMs) and Hebbian Learning to propose the building blocks of Convolutional Self-Organizing Neural Networks (CSNNs), which learn representations in an unsupervised and Backpropagation-free manner. Our approach replaces the learning of traditional convolutional layers from CNNs with the competitive learning procedure of SOMs and simultaneously learns local masks between those layers with separate Hebbian-like learning rules to overcome the problem of disentangling factors of variation when filters are learned through clustering. We investigate the learned representation by designing two simple models with our building blocks, achieving comparable performance to many methods which use Backpropagation, while we reach comparable performance on Cifar10 and give baseline performances on Cifar100, Tiny ImageNet and a small subset of ImageNet for Backpropagation-free methods. 
\end{abstract}
	
	
	\section{Introduction}
	A well-known downside of many successful Deep Learning approaches like Convolutional Neural Networks (CNNs) is their need for big, labeled datasets. Obtaining these datasets can be very costly and time-consuming or the required quantity is even not available due to restrictive conditions. In fact the availability of big datasets and the computational power to process this information are two of the main reasons for the success of Deep Learning techniques \cite{NIPS2012_4824}. This raises the question of why models like CNNs need that much data: Is it the huge amount of model parameters, are training techniques like Backpropagation \cite{werbos,Chauvin:1995:BTA:201784} just too inefficient or is the cause a combination of both?
	
	The fact that neural networks have become both large and deep in recent years, and often consist of millions of parameters, suggests that the cause is a combination of both.
	On the one hand, large amounts of data are needed, because there are so many model parameters to set. On the other hand, the data may not yet be optimally utilized. The suboptimal utilization of data to date could be related to the following problem, which we call the \textit{Scalar Bow Tie Problem}:
	\begin{figure}[tbp]
		\centerline{\includegraphics[width=0.4\columnwidth]{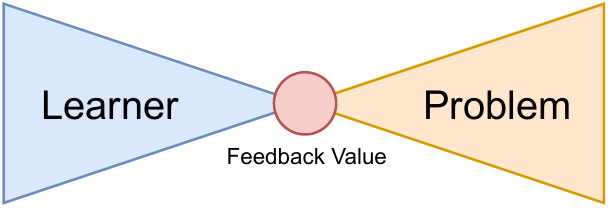}}
		\caption{The Scalar Bow Tie Problem: With this term we describe the current problematic situation in Deep Learning in which most training approaches for neural networks and agents depend only on a single scalar loss or feedback value. This single value is used to subsume the performance of an entire neural network or agent even for very complex problems.}
		\label{scalarbowtie}
	\end{figure}
	Most training strategies for neural networks rely just on one single scalar value (the loss value) which is computed for each training step and then used as a feedback signal for optimizing millions of parameters. In contrast, humans use multiple feedback types to learn, e.g., when they learn representations of objects, different sensor modalities are used. With the Scalar Bow Tie Problem we refer to the current situation for training neural networks and agents. Similar to a bow tie, both ends of the learning setting (the learner and the task to be learned) are complex and the only connection between these complex parts is one single point (the scalar loss or feedback value).
	
	There exists a current effort of Backpropagation-based methods to overcome the Scalar Bow Tie Problem, e.g., by using multiple loss functions at different output locations of the network (a variant of multi-task learning, i.a. \cite{DBLP:journals/corr/abs-1708-07860}). Furthermore, there exists a current debate to which degree Backpropagation is biologically plausible on a network level (e.g., \cite{Lillicrap2016, whittington, DBLP:journals/corr/BengioLBL15}), which often highlights that layer-dependent weight updates in the whole model are less likely than more independent updates, like layer-wise ones, and beyond that would require an exact symmetric copy $W^T$ of the upstream weight $W$. In \cite{GROSSBERG198723} Grossberg describes this problem in detail and refers to it as the \textit{weight transport problem}.
	
	In this work we do not aim to solve all these problems, but want to make a small step away from common Deep Learning methods while keeping some benefits of these models, like modularity and hierarchical structure. We propose a CNN variant with building blocks that learn in an unsupervised, self-organizing and Backpropagation-free manner (CSNN). Within these blocks we combine methods from CNNs \cite{6795724}, Self-Organizing Maps (SOMs) \cite{58325} and Hebbian Learning \cite{10.1007/978-3-642-70911-1_15}.
	
	Our main contributions are as follows:
	\begin{itemize}
		\item We propose an unsupervised, Backpropagation-free learning algorithm, which uses two learning rules to update weights layer-wise without an explicitly defined loss function and thereby reduces the Scalar Bow Tie Problem.
		\item This learning algorithm is used to train CSNN models, with which we achieve comparable performance to many models trained in an unsupervised manner.  
		\item We overcome a fundamental problem of SOMs trained on image patches, by presenting two types of weight masks, to mask input and neuron activities.
		\item We propose a multi-headed version of our building blocks, to further improve performance.
		\item To the best of our knowledge, our approach is the first to combine the successful principles of CNNs, SOMs and Hebbian Learning into a single Deep Learning model.
	\end{itemize}
	
	\section{Related Work}
	Our work relates to methods of unsupervised representation learning. Due to the usage of SOMs, it further relates to clustering methods \cite{saxena}; in fact one could change the SOM learning algorithm with another clustering algorithm. Moreover, we can relate our proposed local learning rules in Section \ref{masklayer} to other local learning rules (e.g., \cite{10.1007/978-3-642-70911-1_15, SANGER1989459, Oja1982}). To stay on track, we focus on methods for unsupervised, Backpropagation(-free) representation learning as follows:

	\subsection{Unsupervised Representation Learning via Backpropagation}
	Recently unsupervised representation learning through auto-encoders \cite{DBLP:journals/corr/abs-1901-04596, DBLP:journals/corr/abs-1812-05069, DBLP:journals/corr/abs-1806-02199, DBLP:journals/corr/abs-1903-05136}, and in general self-supervised approaches \cite{DBLP:journals/corr/abs-1901-09005}, showed promising results. Zhang et al. \cite{DBLP:journals/corr/abs-1901-04596} improved the approach of Gidaris et al. \cite{DBLP:journals/corr/abs-1803-07728} by not only predicting rotations, but by auto-encoding a variety of transformations between images. In \cite{DBLP:journals/corr/abs-1903-05136} a hierarchical, disentangled object representation and a dynamics model for object parts is learned from unlabeled videos by a novel formulation, which can be seen as a conditional, variational auto-encoder with additional motion encoder. Hjelm et al. \cite{hjelm2018learning} proposed \textit{Deep InfoMax}, which defines an objective function that maximizes mutual information between global features created from smaller, random crops of local features and the full sets of local features, while matching a prior distribution to capture desired characteristics like independence. In \cite{DBLP:journals/corr/abs-1904-11567} an objective is defined to search neighborhoods with high class consistency anchored to individual training samples, considering the high appearance similarity among the samples via a softmax function. The results in \cite{DBLP:journals/corr/abs-1906-03248} support our view of the learning problem - the Scalar Bow Tie Problem - since a concept is introduced which uses an evolutionary algorithm (AutoML) to obtain a better weighting between individual losses in a multi-modal, multi-task loss setting for self-supervised learning. Yang et al. \cite{yang2019patch} learns patch features by a discriminative loss and an image-level feature learning loss. The discriminative loss is designed to pull similar patches together and vice versa. The image-level loss pulls the real sample and the surrogate positive samples (random transformations) together while pushing hard negative samples away and serves as an image-level guidance for patch feature generation. To learn representations with minimal description lengths Bizopoulos et al. \cite{DBLP:journals/corr/abs-1907-06592} used i.a. sparse activation functions and minimizes the reconstruction loss dependent on the input and the summed up reconstructions of each individual weight used to reconstruct the entire input. In \cite{DBLP:journals/corr/abs-1904-03436} a Siamese Network with a novel instance feature-based softmax embedding method is used to learn data augmentation invariant and instance spread-out features, by considering this problem as a binary classification problem via maximum likelihood estimation. Thereby only the augmented sample and no other instances should be classified as the unaugmented instance. Donahue et al. \cite{DBLP:journals/corr/abs-1907.02544} showed that GANs can reach state-of-the-art performance by training a \textit{BigBiGAN} on unlabeled ImageNet, and afterwards learning a linear classifier with labels on the latent variable predicted by the encoder of the BigBiGAN, given the input data.
	
	\subsection{Unsupervised, Backpropagation-Free Representation Learning}
	While there are many advances in unsupervised methods using Backpropagation, in comparison there are only a few efforts to propose Backpropagation-free alternatives. Chan et al. \cite{DBLP:journals/corr/ChanJGLZM14} 
	used Principal Component Analysis (PCA) in a convolutional manner to learn two-stage filter banks, which are followed by binary hashing and block histograms to compress the representation. In a similar way Hankins et al. \cite{10.1007/978-3-030-03493-1_87} used SOMs and additionally introduced pooling layers to shrink SOM-activations along the feature dimension. Coates et al. \cite{pmlr-v15-coates11a} studied hyperparameter choices and preprocessing strategies on several single-layer networks learned on convolutional patches, including sparse auto-encoders, sparse restricted Boltzmann machines, K-means clustering, and Gaussian mixture models and showed, e.g., general performance gains when the representation size is increased. In \cite{4469953} an adaptive-subspace SOM is used for saliency-based, invariant representation learning for image classification. Krotova et al. \cite{Krotov7723} introduced a learning algorithm motivated by Hebbian Learning to learn early feature detectors by utilizing global inhibition in the hidden layer. In \cite{pmlr-v32-line14} a Nonnegativity Orthogonal Matching Pursuit (NOMP) encodes representations by selecting a small number of atoms from the feature dictionary, such that their linear combination approximates the input data vector best under an nonnegativity constraint. This constraint mitigates OMP’s stability issue and adds resistance to overfitting. Furthermore, there exists work on unsupervised spiking neural network models using spike-timing dependent plasticity learning rules (e.g., \cite{DBLP:journals/corr/abs-1904-06269}). 
	
	\section{CSNN}
\begin{figure}[tbp]
		\centerline{\includegraphics[width=0.75\columnwidth]{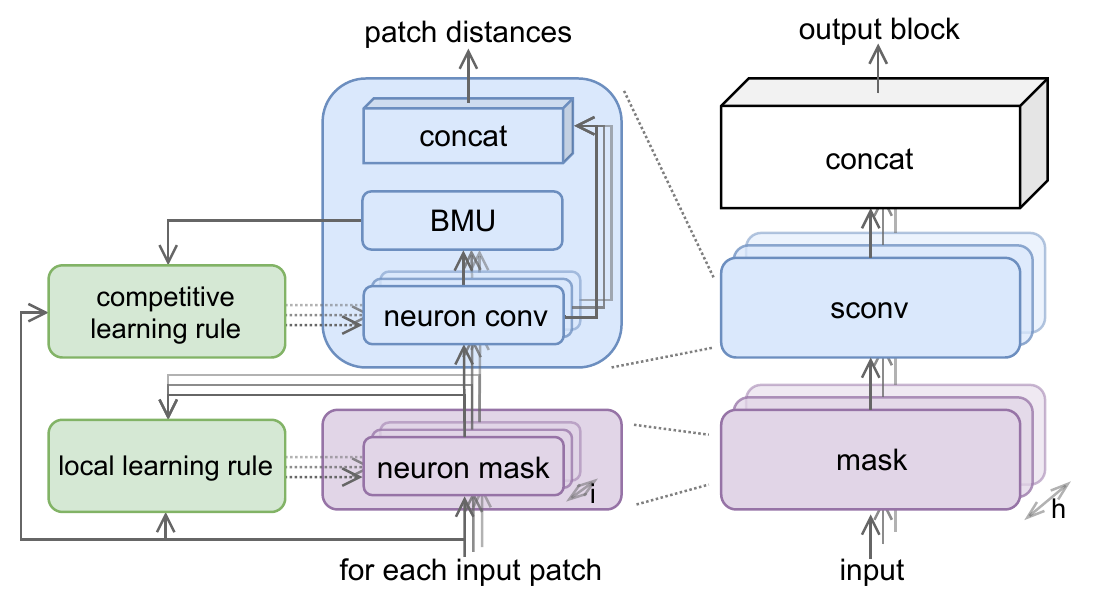}}
		\caption{(left) mask and sconv layer computations for a single patch: First each patch is masked for each neuron (i). The masked outputs and the input patch will be used in the local learning process. Then the sconv layer computes the convolution distances and the BMU. The BMU and the input patch will be used in competitive learning. The distances are concatenated to the patchs spatial activation vector in the convolution output block. (right) Multi-Head masked sconv: Multiple mask and sconv layers (h) are applied on the entire input and the resulting output blocks are concatenated in the feature dimension.}
		\label{keyblocks}
	\end{figure}

	This section describes the key methods of CSNNs. The proposed blocks for each method and their interactions are summarized in Fig. \ref{keyblocks}.
	
	\subsection{Convolutional, self-organizing Layer}
	The convolutional, self-organzing layer (sconv) follows closely the convolutional layer of standard CNNs. The entire input is convolved patchwise with each learned filter:
	\begin{equation}
	y_{m,n,i}=\mathbf{p}_{m,n}\cdot \mathbf{w}_{i}
	\label{eqn:conv}
	\end{equation}
	where $\mathbf{p}_{m,n}$ is the image patch centered at position $m, n$ of the input, $\mathbf{w}_i$ is the $i$-th filter of the layer and $y_{m,n,i}$ is the activation at position $m,n,i$ of the output. The only difference to a standard convolutional layer is that the weights are learned by a competitive, self-organizing optimization procedure. The output of the layer for one image patch are the distances for each neuron in a SOM. Each neuron of the SOM corresponds to one filter in a CNN, and each distance output of a SOM-neuron corresponds to a specific feature $y_{m,n,i}$ of the feature map $i$ at the center of the patch $m,n$ after the convolution operation was applied. The central difference to a standard CNN is that the CSNN only uses local learning rules. It is trained unsupervised and bottom-up fashion in contrast to the top-down supervised backpropagation approach.
	Analog to CNNs the proposed modular layers can be combined with other layers in Deep Learning architectures.
	
	\subsection{Competitive Learning Rule}
	Most learning rules for SOMs require a best matching unit (BMU) in order to compute the weight change $\Delta w$. Since we use a convolution as distance metric the index of the BMU for one patch is defined as:
	\newcommand{\argmax}{\mathop{\mathrm{argmax}}}
	\begin{equation}
	c_{m,n}=\argmax_{i}\{y_{m,n,i}\}
	\label{eqn:BMU}
	\end{equation}
	where $c_{m,n}$ is the index of the BMU in the (2D) SOM-grid for patch $m,n$. 
	
	To allow self-organization, learning rules for SOMs require a neighborhood function to compute the neighborhood coefficients from all the other neurons in the grid to the BMU. A common type of neighborhood function is the Gaussian:
	\begin{equation}
	h_{m, n, i}(t)=\exp(\frac{-d_{A}(k_{m,n,c_{m, n}}, k_{m,n,i})^2}{2\delta(t)^2})
	\label{eqn:nfunction}
	\end{equation}
	where $k_{m,n,c_{m, n}}$ and $k_{m,n,i}$ are the coordinates of the BMU and the $i$-th neuron in the SOM-grid. These coordinates depend on the dimensionality of the grid. For the distance function $d_A$ we use the Euclidean distance and $\delta(t)$ is a hyperparameter to control the radius of the Gaussian which could change with the training step ($t$). Then $h_{m, n, i}(t)$ is the neighborhood coefficient of neuron $i$ to the center (the BMU) $c_{m,n}$ for the patch $m,n$. In our experiments we did not investigate other neighborhood functions.
	
	Now the weight update for one patch can be defined as:
	\begin{equation}
	\Delta \mathbf{w}_{m,n,i}(t)=a(t)h_{m, n, i}(t)\mathbf{p}_{m,n}
	\end{equation}
	\begin{equation}
	\mathbf{w}_{m,n,i}(t)=\frac{\mathbf{w}_{m,n,i}(t)+\Delta \mathbf{w}_{m,n,i}(t)}{\left\| \mathbf{w}_{m,n,i}(t)+\Delta \mathbf{w}_{m,n,i}(t)\right\|}
	\end{equation}
	where $\left\|...\right\|$ is the Euclidean norm used to get the positive effects of normalization such as preventing the weights from growing too much. $a(t)$ is the learning rate which could change over the course of the training. \\
	For an entire image, the final weight change for the weight vector of a SOM-neuron is computed by taking the mean over all patch weight updates:
	\begin{equation}
	\Delta \mathbf{w}_{i}(t)=\frac{a(t)}{mn}\sum_{m,n}h_{m, n, i}(t)\mathbf{p}_{m,n} 
	\label{eqn:slearning}
	\end{equation}
	In batch training the mean of all patches in a batch is taken.
	
	The use of other distance metrics (like the L1- or L2-norm) may make it necessary to switch the computation of the BMU (e.g., to argmin) and the learning rule.
	
	\subsection{Mask Layers}\label{masklayer}
	We argue that given the learning rule \eqref{eqn:slearning} a SOM-neuron is not able to disentangle factors of variation, which is a key property of Deep Learning architectures \cite{Goodfellow-et-al-2016}. In other words, it is not able to choose which part of the input it wants to give more attention and which it wants to ignore. The SOM just tries to push its neurons around to fit the dataset best. This may lead to poor performance, because a neuron is not able to learn abstract features which can be seen as concepts. Even in higher layers a SOM-neuron just processes the input from a collection of SOM-neurons of the previous layer in its receptive field. Therefore, the network forms a hierarchy of collections rather than a hierarchy of concepts. Furthermore, a neuron in a deeper layer with a small receptive field of e.g., $3\times3$ needs to compute the distance between its $3\times3\times256$ vector and the input patch, if there are $256$ SOM-neurons in the previous layer. Because the weight update “shifts” all the values of the weights from the BMU and its neighbors in one direction to the input at once, it seems unlikely that the SOM-neurons learn distinct enough features for data with a high amount of information such as images. This argument goes hand in hand with Dundar et al. \cite{DBLP:journals/corr/DundarJC15}, where similar observations are made for k-means clustering.
	
\begin{figure}[tbp]
		\centerline{\includegraphics[width=0.95\columnwidth]{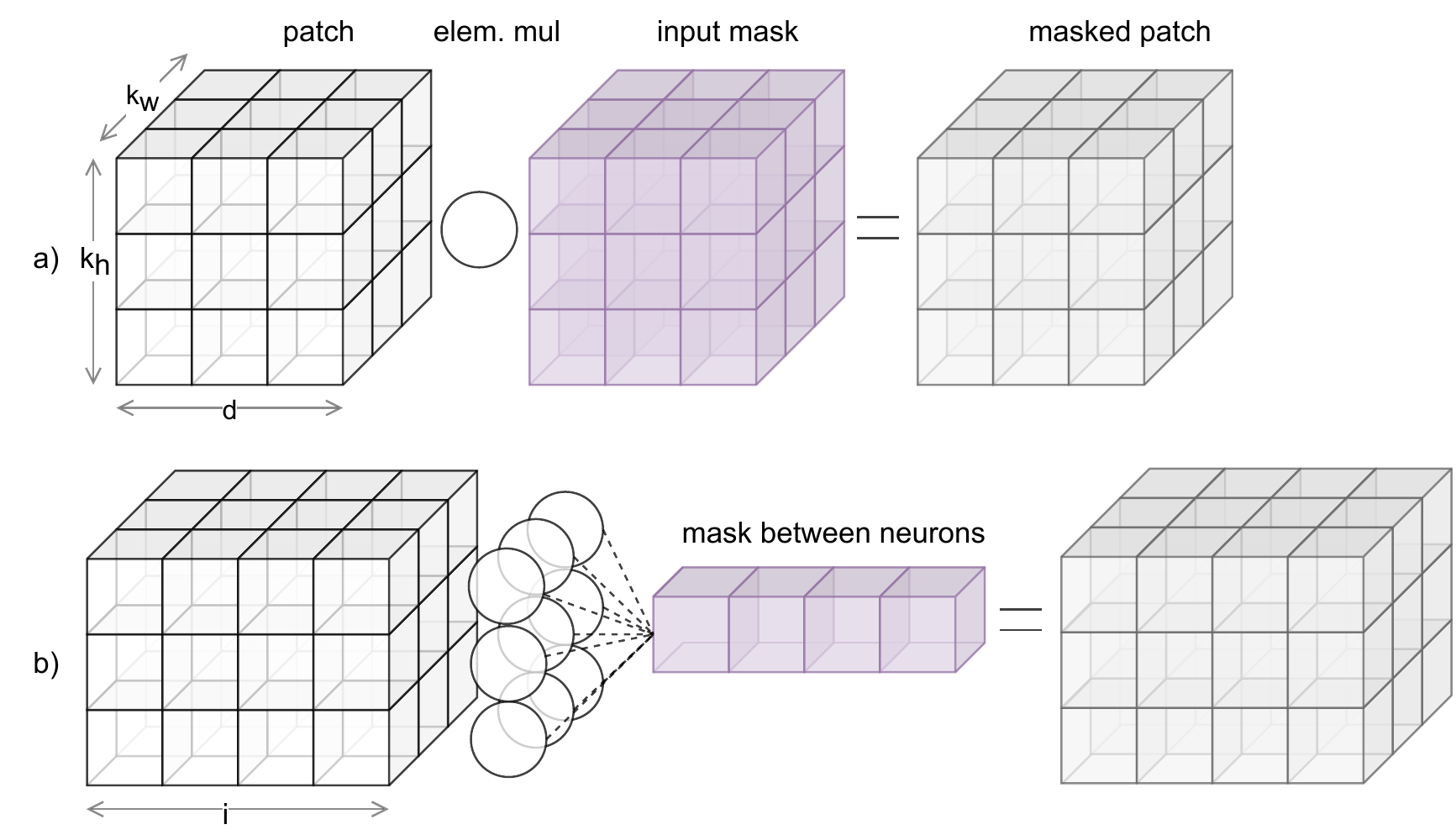}}
		\caption{a) Input mask: The Hadamard product between an input patch and a mask of same size. b) Mask between layers: The mask, with its length equal to the number of filters in the previous layer, is multiplied element-wise with every spatial activation vector from the patch.}
		\label{masktyps}
	\end{figure}

	To allow the network to learn some kind of concept, we propose two types of separately learned mask vectors shown in Fig. \ref{masktyps}. The first mask (a) enables each neuron to mask its input values with a vector of the same dimension as the input patch and the SOM-neuron's weight vector. Each CSNN-neuron multiplies the image patch with its mask and convolves the masked patch with the weight vector of the SOM neuron:
	\begin{equation}
	\mathbf{\hat{y}}_{m,n,i}=\mathbf{p}_{m,n}\circ \mathbf{m}_{i}
	\label{eqn:inputmask}
	\end{equation}
	\begin{equation}
	y_{m,n,i}=\mathbf{\hat{y}}_{m,n,i}\cdot \mathbf{w}_{i}
	\label{eqn:maskpatch}
	\end{equation}
	where $\mathbf{m}_{i}$ is the mask of neuron $i$, $\mathbf{\hat{y}}_{m,n,i}$ is the masked patch for neuron $i$ and $\circ$ denotes the Hadamard product.
	
	The second mask (b) lies between the neurons of two layers. Therefore, it requires $k_w\times k_h$ fewer parameters then the first (input) mask, when $k_w$ and $k_h$ are the kernel's sizes. The mask is defined as:
	\begin{equation}
	\mathbf{\hat{y}}_{m_h,n_w,i}=\mathbf{p}_{m_h,n_w}\circ \mathbf{n}_{i}
	\label{eqn:neuronmask}
	\end{equation}
	where $\mathbf{n}_{i}$ is the mask element-wise multiplied in depth $k_w\times k_h$ times. With this mask we want to enable each neuron to choose from which neuron in the previous layer it wants to get information. Mask (a) is used for the input, and mask (b) between each sconv layer.
	\\
	\\
	We note that we still use the unmasked input for the training of the SOM-weight vectors — the mask is only used to compute the convolution distances and to determine the BMU. Moving the BMU and its neighbors into unmasked input direction leads to a more stable learning process, since the self-organization does not rely to heavily on the higher mask coefficients. High mask coefficients could drive the weight vector of a single SOM-neuron into a direction from which it can hardly adapt to other new, but similar input patches. Thereby the neuron could get stuck in a small dead end. Furthermore, this gives the mask the opportunity to learn new regions typical for a particular input type even when the mask tends to ignore these regions of the input. This is because the SOM moves in the direction of the unmasked input, which opens the opportunity for the mask to increase its corresponding coefficients.
	
	\subsection{Local Learning Rules}
	To learn the input mask of each neuron, we propose two simple local learning rules which are inspired by the Generalized Hebbian Algorithm \cite{SANGER1989459} (GHA). These procedures show good results in our experiments, but there are certainly other local learning rule alternatives that could guide the training process of the masks as well.
	
	Below we derive our final learning rules by discussing the following equations within the context of the CSNN, where $e$ indicates that we present the element-wise form of the learning rules. For the original inviolate versions we refer to the literature.
	\begin{equation}
	\footnotesize
	\small
	\Delta h_e(\mathbf{p}_{m, n}, \mathbf{\hat{y}}_{m, n, i}) = \mathbf{\hat{y}}_{m, n, i} \circ \mathbf{p}_{m, n} 
	\label{eqn:simple_hebbian_learning}
	\end{equation}
	\begin{equation}
	\footnotesize
	\small
	\Delta o_e(\mathbf{p}_{m, n}, \mathbf{\hat{y}}_{m, n, i}, \mathbf{m}_{i}) = \mathbf{\hat{y}}_{m, n, i} \circ (\mathbf{p}_{m, n} - \mathbf{\hat{y}}_{m, n, i} \circ \mathbf{m}_{i})
	\label{eqn:ojas_rule}
	\end{equation}
	\begin{equation}
	\footnotesize
	\Delta m_{es}(\mathbf{p}_{m, n}, \mathbf{\hat{y}}_{m, n, i}, \mathbf{m}_{i}) = h_e([\mathbf{p}_{m, n}-\gamma\sum_{k}\mathbf{\hat{y}}_{m,n,k}\circ \mathbf{m}_{k}], \mathbf{\hat{y}}_{m, n, i}, \mathbf{m}_{i})
	\label{eqn:similar_to_gha}
	\end{equation}
	\begin{equation}
	\footnotesize
	\Delta m_{ec}(\mathbf{p}_{m, n}, \mathbf{\hat{y}}_{m, n, i}, \mathbf{m}_{i}) = o_e([\mathbf{p}_{m, n}-\gamma\sum_{k<i}\mathbf{\hat{y}}_{m,n,k}\circ \mathbf{m}_{k}], \mathbf{\hat{y}}_{m, n, i}, \mathbf{m}_{i})
	\label{eqn:similar_to_gha_with_restricted_summation}
	\end{equation}
	\begin{equation}
	\footnotesize
	\Delta \mathbf{m}_{i}(t)=\frac{a(t)}{mn}\sum_{m,n}[\Delta m_{es}(\mathbf{p}_{m, n}, \mathbf{\hat{y}}_{m, n, i}, \mathbf{m}_{i})]
	\label{eqn:final_update_formula_for_input_masks}
	\end{equation}
	\begin{equation}
	\footnotesize
	\Delta \mathbf{n}_{i}(t)=\frac{a(t)}{mnhw}\sum_{m,n, h, w}[\Delta m_{es}(\mathbf{p}_{m_h,n_w}, \mathbf{\hat{y}}_{m_h,n_w,i}, \mathbf{n}_{i})]
	\label{eqn:final_update_formula_for_masks_between_neurons}
	\end{equation}
	Eq. \eqref{eqn:simple_hebbian_learning} describes simple Hebbian Learning in vector notation. It implements the principle ``neurons that fire together, wire together'', where $\mathbf{p}_{m, n}$ is the pre- and $\mathbf{\hat{y}}_{m, n, i}$ is the postsynaptic neuron activities vector. This principle per se is useful for mask learning, because the procedure could find mask values, which indicate the connection strengths between mask input and output. However, Hebb's rule is unstable. If there are any dominate signals in the network — and in our case this is likely due to BMUs — the mask values will approach numerical positive or negative infinity rapidly. In fact, it can be shown that the instability of Hebbian Learning accounts for every neuron model \cite{Principe:1999:NAS:555238}.
	
	Eq. \eqref{eqn:ojas_rule}, known as Oja's rule \cite{Oja1982}, tries to prevent this problem through multiplicative normalization, where the resulting additional minus term can be seen as the forgetting term to control the weight growth. This restricts the magnitude of weights to lie between 0 and 1.0, where the squared sum of weights tends to 1.0 in a single-neuron fully-connected (fc) network \cite{Oja1982}. For that reason, using Oja's rule in our case would lead to mask values approaching 1.0, because each mask value has its own input and output due to the element-wise multiplication and is therefore independent of all other mask value updates — each mask value forms a one-weight, single-neuron fc network. In order to prevent this tendency to 1.0, we can make the mask value updates dependent on each other. 
	
	The GHA \cite{SANGER1989459} makes its updates dependent through input modification to approximate eigenvectors for networks with more than one output allowing to perform a neural PCA \cite{SANGER1989459} in contrast to Oja's rule, which approximates the first eigenvector for networks with only one output. In \cite{SANGER1989459} one can see that the algorithm is equivalent to performing Oja Learning using a modified version of the input, and a particular weight is dependent on the training of the other weights only through the modifications of the input. If Oja's algorithm is applied to the modified input, it causes the $i-$th output to learn the $i$-th eigenvector \cite{SANGER1989459}. However, our mask values have their own single and independent inputs and outputs.
	
	Therefore, in \eqref{eqn:similar_to_gha} we use a modified input patch for mask $i$ by simply subtracting the input patch from the sum over mask outputs, which sums the filtered information of all masks when $k$ iterates over all masks. Similar to the GHA we additionally multiply the mask to the output before summing up, which leads to a normalization effect and drives the mask coefficients to the initialization interval $[-1, 1]$. Now each mask update tries to incorporate the missing input information in the output in a competitive manner, where the mask value growth is restricted. This can be seen as a self-organization process for the masks. In our experiments we found that due to the summation over all masks the updates are stable enough to use Hebbian Learning \eqref{eqn:simple_hebbian_learning}, which saves computation time.
	
	In \eqref{eqn:similar_to_gha_with_restricted_summation} we sum up all $k<i$ and each next mask $\mathbf{m}_i$ sees an output from which the filtered information of the previous masks $\{\mathbf{m}_{1},...,\mathbf{m}_{i-1}\}$ in the grid is subtracted. The hyperparameter $\gamma$ is used to control how much information we want to subtract from the input. A smaller $\gamma < 1.0$ value enables the masks to share more input information with each other. 
	
	Eq. \eqref{eqn:final_update_formula_for_input_masks} shows the final update formula for the input masks \eqref{eqn:inputmask}, where we compute the mean over every patch $\mathbf{p}_{m,n}$. For the second mask type \eqref{eqn:neuronmask} the final update formula is shown in \eqref{eqn:final_update_formula_for_masks_between_neurons}. Instead we update here the mask $\mathbf{n}_i$ with the mean over each spatial activation vector $\mathbf{p}_{m_h,n_w}$ of every patch. Experiments show that the update rules \eqref{eqn:final_update_formula_for_input_masks} and \eqref{eqn:final_update_formula_for_masks_between_neurons} lead to barley better performance for smaller models when using the input modification of \eqref{eqn:similar_to_gha_with_restricted_summation} and $\gamma=0.5$. But this equation requires the Oja Rule and sometimes bigger neighborhood coefficients — especially when $\gamma$ is near $1.0$ — because of unequal opportunities of masks to learn which may lead to poorer efficiency. With the other input modification \eqref{eqn:similar_to_gha} and $\gamma=1.0$ we achieve better performance for our deeper models.
	
	Furthermore, we want to note that in our experiments the mask values are initialized uniformly in $[-1,1]$ to encourage the model to learn negative mask values, which allow the masks to flip their input. For stability it is recommended that the input is normalized. It is also possible to multiply $\mathbf{m}_{i}$ by $h_{m, n, i}(t)$ to get similar masks for neighboring neurons, but experiments show that the resulting loss in diversity decreases performance. Therefore we only update the BMU mask.
	
	\subsection{Multi-Head Masked SConv-Layer}
	To further improve the performance of our models we do not use one big SOM per layer, but multiple smaller maps. For the output of a layer we concatenate the SOMs outputs in the feature dimension. We use multiple maps for multiple reasons: First, the neuron's mask may (and should) lead to specialization, and in combination with the self-organization process a single SOM is possibly still not able to learn enough distinct concepts. Second, multiple smaller maps seem to learn more distinct features more efficiently than scaling up a single map.
	This approach is very flexible, in fact, experiments show similar performance between SOMs using $3$ big maps or $12$ small maps per layer. To further increase diversity we update only the best BMU of all maps per patch. This could lead to maps that will never be updated or to maps where the update is stopped to early during training. We do not use a procedure to prevent these dead maps, because we have not had problems with these during our experiments. In fact, dead maps can often be prevented by scaling up the coefficient of the neighborhood function \eqref{eqn:nfunction}. It is a task for future research to examine dead maps and to investigate whether the use of multiple SOMs in the CSNN model leads to redundant information. 
	\\
	\\
	All the presented learning methods can be applied layer-wise. On the one hand, this brings benefits like the opportunity to learn big models by computing layer by layer on a GPU and more independent weight updates compared to gradient descent, which reduces the Scalar Bow Tie Problem. On the other hand, this independence could lead to fewer ``paths'' through the network, reducing the capability of the network to disentangle factors of variation. Examining this problem by creating and comparing layer-wise dependent and independent learning rules is an interesting direction for future research, and some work has already been done (e.g., \cite{Lillicrap2016}).
	
	\subsection{Other Layers and Methods}
	Since we are in the regime of convolutional Deep Learning architectures, there are plenty of other methods we could test within the context of our modules. In our experiments we choose to investigate batch normalization \cite{DBLP:journals/corr/IoffeS15} (with no trainable parameters) and max-pooling \cite{NIPS2012_4824}. Furthermore, SOMs are pretty well studied and there are many improvements over the standard formulation of the SOM, which could be investigated in the future (e.g., the influence of different distance metrics).
	
	\section{Experiments}
	To evaluate our methods we design and follow two model architectures throughout the experiments. For training our CSNNs we do not use any fancy strategies like pre-training, decaying rates or regularization. We simply use the zero-mean and unit-variance of the datasets. Our implementation is in TensorFlow and Keras and available at \href{https://github.com/BonifazStuhr/CSNN}{https://github.com/BonifazStuhr/CSNN}.
	
	\subsection{Datasets}
	\textbf{Cifar10} \cite{Krizhevsky2009LearningML} with its $32\times32$ images and 10 classes is used to evaluate the importance of the proposed building blocks. We split the validation set into $5000$ evaluation and test samples.
	
	\textbf{Cifar100} \cite{Krizhevsky2009LearningML} contains 100 classes with which we test the capability of our representation to capture multiple classes. We split the validation set into $5000$ evaluation and test samples.
	
	\textbf{Tiny ImageNet} is used to investigate the capability of our models to learn on a dataset with larger images ($64\times64$) containing 200 classes with fewer examples.
	
	\textbf{SOMe ImageNet}: We take a subset of ImageNet \cite{5206848} containing $10100$ training, $1700$ evaluation and $1700$ test samples and 10 classes to test our performance on larger ($256\times256$) images. Please refer to our implementation for details.
	
	\subsection{Evaluation Metrics}
	To evaluate our models we mainly use the representations to train a \textbf{linear (l)}, a \textbf{nonlinear (nl)} and a \textbf{few-shot (fs) classifier}. To further evaluate the model we compute the percentage of used neurons per batch (\textbf{neuron utilization}), try to \textbf{reconstruct the input} from the representation and have a look at different \textbf{neuron activities} based on their input class. We prove the stability of our learning processes by reporting \textbf{10-fold cross-validation} results. Furthermore, we give \textbf{random (R) baseline} performances for our models. At this point we want to highlight the importance of such baselines, e.g., Saxe et al. \cite{Saxe10onrandom} showed that models with random weights can achieve performances often just a few percent worse than sophisticated unsupervised training methods. Additional plots and visualizations can be found in the Appendix \ref{appendixvis}.
	
	\subsection{Model Architectures and Details}
	Appendix \ref{appendixvis}, Fig. \ref{csnnarchitecture} shows a CSNN with three maps.
	
	\textbf{S-CSNN}: Our small model consists of two CSNN-layers, each is followed by a batch norm and a max-pooling layer to halve the spatial dimension. The first layer uses a $10\times10\times1$ SOM-grid (1 head with 100 SOM-neurons), stride $2\times2$ and input masks \eqref{eqn:final_update_formula_for_input_masks} the second layer a $16\times16\times1$ SOM-grid, stride $1\times1$ and masks between neurons \eqref{eqn:final_update_formula_for_masks_between_neurons}. Both layers use a kernel size of $3\times3$ and the padding type ``same''.
	
	\textbf{D-CSNN}: Our deeper model consists of three CSNN-layers, each is followed by a batch norm and a max-pooling layer to halve the spatial dimension. The first layer uses a $12\times 12\times 3$ SOM-grid and input masks \eqref{eqn:final_update_formula_for_input_masks}, all remaining layers use masks between neurons \eqref{eqn:final_update_formula_for_masks_between_neurons}. The second layer consist of a $14\times 14\times 3$ SOM-grid, and the third layer of a $16\times 16\times 3$ SOM-grid. All layers use a kernel size of $3\times3$, a stride of $1\times1$ and the padding type ``same''. For Tiny ImageNet the layer 1 stride is $2\times2$ and for SOMe ImageNet the layer 1 and 2 strides are $3\times3$ to keep the representation sizes the same. 
	
	\textbf{Training}: We simply set the learning rate of the SOM(s) to $0.1$ and the learning rate of the local mask weights to $0.005$ for all layers. The neighborhood coefficients for each layer are set to $(1.0,1.25)$ for the S-CSNN and $(1.0,1.5,1.5)$ for the D-CSNN. The individual layers are learned bottom up, one after another, because its easier for deeper layers to learn, when the representations of the previous layers are fully learned. The steps for which each layer will be learned can be defined in a flexible training interval. 
	
	\textbf{Classifiers}: Our nl classifier is a three layer MLP (512, 256, num-classes) with batch norm and dropout $(0.5,0.3)$ between its layers. Our l and fs classifier is simply a fc layer (num-classes). All classifiers use elu activation functions and are trained using the Adam optimizer with standard parameters and batch size $512$. To infer the results on the test set we take the training step where the evaluation accuracy of the model was the best (pocket algorithm approach).
	
	\subsection{Objective Function Mismatch}
\begin{figure}[tbp]
		\centerline{\includegraphics[width=1.0\columnwidth]{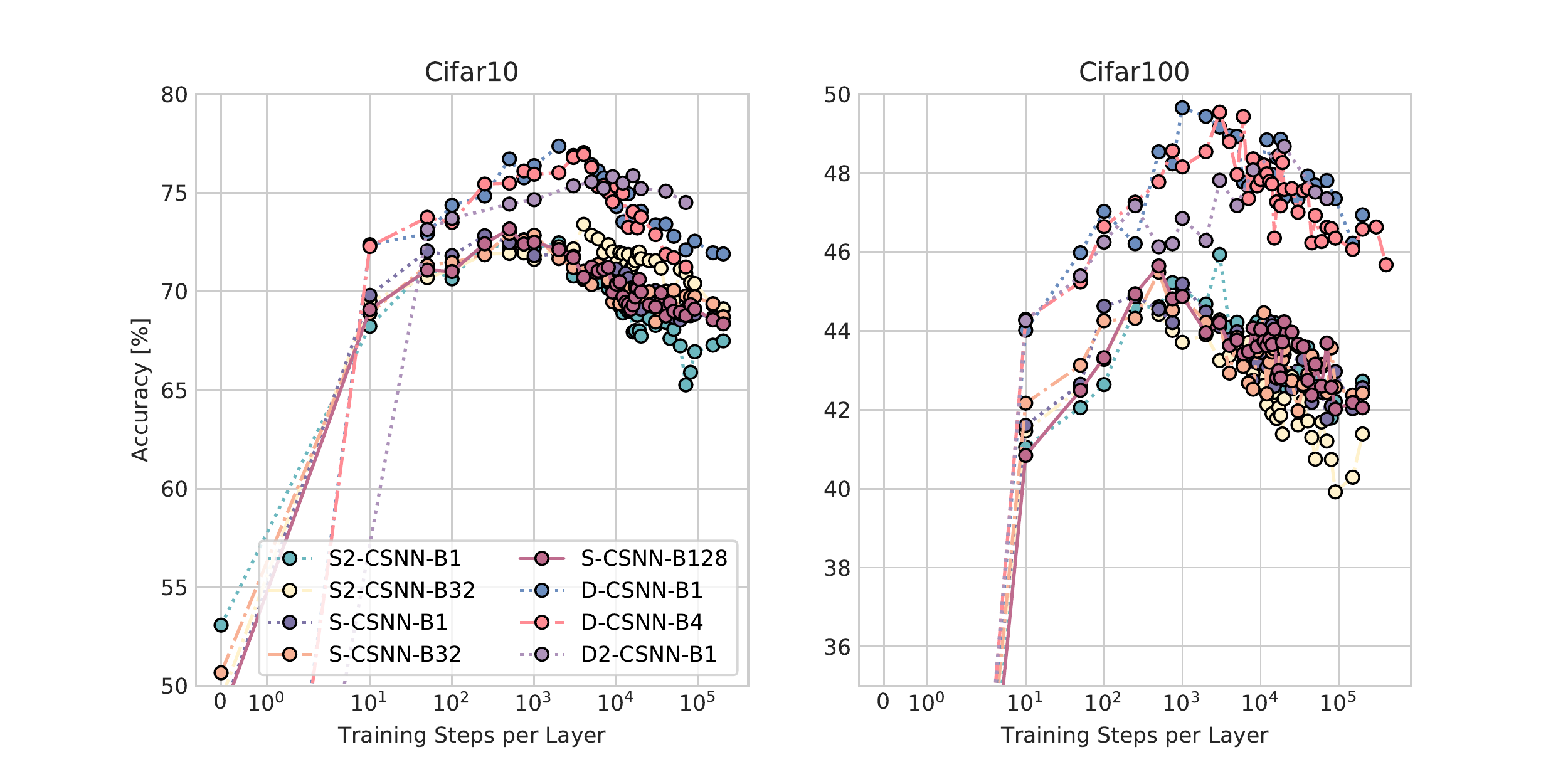}}
		\caption{Accuracies of our models trained with our local mask learning rules (S/D for \eqref{eqn:similar_to_gha} and S2/D2 for \eqref{eqn:similar_to_gha_with_restricted_summation}) and different batch sizes B. Every point corresponds to the test dataset accuracy of an nl classifier which was trained on the representation of the CSNN for every sample in the training dataset. The layers of the CSNN where trained layer-wise for the steps shown on the x-axes. Best viewed in color.}
		\label{ofm}
	\end{figure}

	Fig. \ref{ofm} shows that all models reach their accuracy peak very early in training. Because this peak is often reached before seeing the entire training set and Fig. \ref{weightchanges} in Appendix \ref{appendixvis} shows ongoing SOM-weight convergence, it seems likely that the decrease in performance is due to the objective function mismatch between the self-organization process and the ideal objective function to learn features for classification. This is a common problem in unsupervised learning techniques \cite{metz2018learning}. Taking another, positive view, one could argue that our models require fewer samples to learn representations (e.g., 2000 samples ($\approx3\%$) of Cifar10 for the D-CSNN-B1 peak). Even more surprising is the huge jump in accuracy (by up to 62\%) after the models have only seen $10$ examples. This could be a positive effect of reducing the Scalar Bow Tie Problem, but we also want to note that better initialization and not using dropout in the nl classifier leads to better initial performance. Furthermore, we can see that the mask learning rule \eqref{eqn:similar_to_gha} is superior in this setting for the deeper model and vice versa.
	\subsection{Influences of the Methods}
	The left part of Table \ref{evalparameters} shows the influence of our proposed building blocks. We start with our small and deep model with random sconv weights and no masks (RS-NM). Then we incrementally add and train our methods showing that: 1) Training only the SOM (NM) decreases performance. 2) Even static, random masks (RM) or noise masks (NO), which change every training step, lead to a huge accuracy increase. We generated these mask from a uniform distribution in $[-1, 1]$. 3) Training masks (no acronym) increases performance again; even if we leave the SOM untrained (RS), accuracy increases through mask training. 4) Our overall learning increases accuracy by about 10\% for the l classifier and 20-65\% for the nl classifier compared to the uniform (in $[-1, 1]$) random baseline (R). The low accuracy of the nl classifier for random weights is due to the use of dropout.\\
	\indent In the right part of Table \ref{evalparameters} we show that: 1) Increasing the amount of maps (single M) improves performance (1M, 2M, no acronym). 2) Similar performance can be reached if a higher number of smaller maps is used (M-12M: $(6\times6, 7\times7, 8\times8)$). 3) On the other hand, updating neighboring masks (NH) or not using batch normalization (NBN) decreases performance. We note that for deeper models batch normalization is necessary to prevent too large values in deeper layers (exploding weights). 4) Updating all maps per step (BMU) decreases the performance of the nl classifier. 5) Using training data augmentation (rotations, shifts and horizontal flip) only to create a bigger representation dataset for the classifier learning increases performance (Aug). Analog to
	\cite{DBLP:journals/corr/abs-1901-09005} we observe that increasing the representation size improves performance, but in contrast classifier training takes not long to converge.
	\begin{table}[tbp]
		\tiny
		\caption{Influences of the Methods with D-CSNN-B1 and S2-CSNN-B32}
		\begin{center}
			\begin{tabular}{|l|c|c|l|c|c|}
				\hline
				\textbf{}&\multicolumn{2}{|c|}{\textbf{Cifar10}}&	\textbf{}&\multicolumn{2}{|c|}{\textbf{Cifar10}} \\
				
				\cline{2-3}
				\cline{5-6}
				
				\textbf{Model} & \textbf{\textit{l}}& \textbf{\textit{nl}}&
				\textbf{Model} & \textbf{\textit{l}}& \textbf{\textit{nl}}\\					
				\hline
				S2-RSNM	&$54.64$    &$15.32$     &  D &$72.66\substack{+0.40 \\ -1.07}$ &$\mathbf{77.21\substack{+0.62 \\ -0.56}}$    \\
				D-RSNM		&$61.86$    &$16.01$     & S &  $\mathbf{66.43\substack{+0.96 \\ -1.02}}$ &$72.79\substack{+0.59 \\ -0.84}$\\
				S2-NM	&$40.37$    &$22.32$    &    D2  &$71.74\substack{+0.99 \\ -1.42}$   &$76.18\substack{+0.77 \\ -1.61}$  \\
				D-NM	&$43.33$    &$12.13$      & S2	&$66.18\substack{+1.17 \\ -0.95}$    &$\mathbf{72.82\substack{+1.02 \\ -1.22}}$  \\
				S2-RM	&$62.89$    &$70.42$      & D-1M  &$69.69$    &$71.86$ \\
				D-RM	&$69.86$    &$73.69$     & D-2M&$72.05\substack{+0.67 \\ -0.99}$   &$76.11\substack{+0.49 \\ -0.57}$  \\
				S2-NO	&$56.84$    &$14.83$     &M-12M     &$71.28$     &$75.77$ \\
				D-NO	&$65.53$   &$20.22$    	 &  D-BMU	    &$\mathbf{73.14\substack{+0.73 \\ -0.70}}$     &$76.88\substack{+0.59 \\ -0.90}$  \\
				S2-RS	&$65.20$    &$67.75$    &		D-NH  &$71.70$     &$75.08$   \\
				D-RS	     &$69.69$      &$69.74$   & 		D-NBN   &$71.67$ 	      &$76.16$ \\
				S2-R	&$56.42\substack{+1.17 \\ -1.46}$   &$52.23\substack{+2.16 \\ -2.48}$   &S2-Aug     &$66.32$   &$\mathbf{76.90}$   \\
				D-R	&$63.36\substack{+1.28 \\ -1.34}$     &$13.44\substack{+5.25 \\ -3.22}$    	& D-Aug	& $\mathbf{73.73}\substack{+1.95 \\ -1.17}$   &$\mathbf{80.03\substack{+1.52 \\ -1.05}}$     \\
				\hline
			\end{tabular}
			\label{evalparameters}
		\end{center}
	\end{table}
	\\
	\begin{figure}[tbp]
		\centerline{\includegraphics[width=\columnwidth]{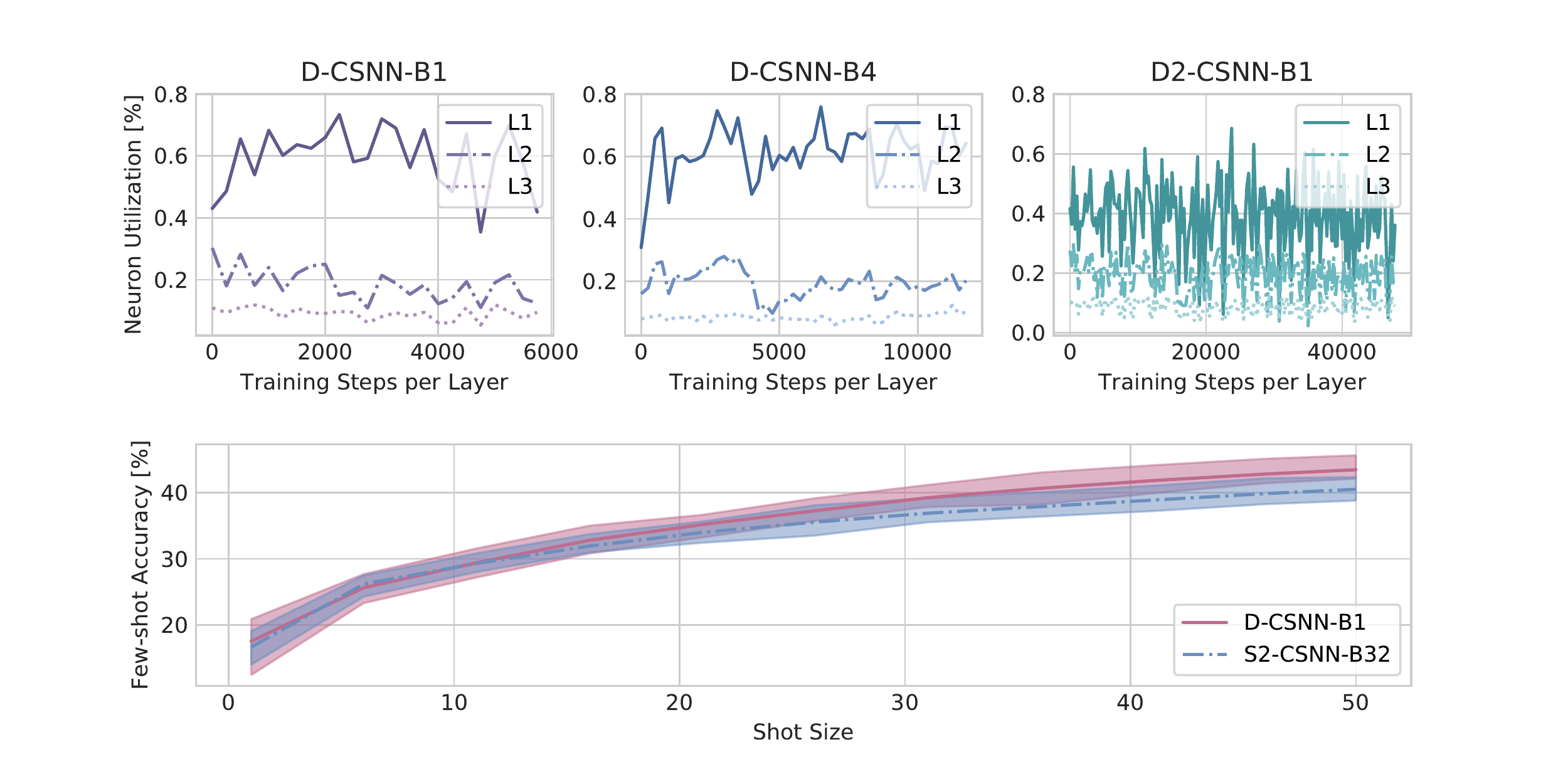}}
		\caption{(top) The neuron utilization of models trained with various batch sizes and mask learning rules on Cifar10. (bottom) The 10-fold few-shot classifier accuracy of these models with increasing shot sizes from $1$ to $50$ on Cifar10.}
		\label{utilAndFs}
	\end{figure}
	The upper part of Fig. \ref{utilAndFs} shows that for every model the neuron utilization — the percentage of used neurons per input — is getting smaller if we move towards deeper layers, indicating that the SOM-neurons of deeper layers specialize in certain inputs. Furthermore, we can see that measuring on a larger batch size leads to slightly higher neuron utilization, indicating that a fair amount of neurons serve a purpose for the dataset and are not dead. In comparison to learning rule \eqref{eqn:similar_to_gha_with_restricted_summation} the rule \eqref{eqn:similar_to_gha} leads to a higher neuron utilization, which makes sense due to the fairer competition. We want to note that a higher neuron utilization does not necessarily lead to a higher performance, because a lower neuron utilization may lead to better distinguishable, sparser representations.

	The lower part of Fig. \ref{utilAndFs} shows how the classification accuracy of the l classifier increases, when the number of representation vectors per class used to learn the l classifier increases, reaching over 41\% when trained on 50 examples per class.

	\subsection{Performances on the Datasets and Generalization}
	Table \ref{performances} shows the performance of our best models compared to other methods. For fs training 50 samples per class were used. The model we use for Cifar100, Tiny ImageNet and SOMe ImageNet is not tuned for these datasets. It uses the same hyperparameters as our Cifar10 model (except the training steps for Cifar100). We can see that D-CSNN-B1 is comparable to many recent methods, but seems to lack performance for Tiny ImageNet, maybe because this amount of classes requires bigger models. To additionally test the generalization capability we train our model on one dataset and test it on the remaining datasets, achieving excellent generalization capability, where difference in accuracy lie in the low percentage range.  
	
\begin{table*}[tbp]
	\tiny
	\caption{Performances of the D-CSNN-B1 model.}
	\begin{center}
		\begin{tabular}{|l|c|c|c|c|c|c|c|c|c|c|c|c|}
			\hline
			\textbf{}&\multicolumn{3}{|c|}{\textbf{Cifar10}} &\multicolumn{3}{|c|}{\textbf{Cifar100}}  &\multicolumn{3}{|c|}{\textbf{Tiny ImageNet}} 
			&\multicolumn{3}{|c|}{\textbf{SOMe ImageNet}} 
			\\
			\cline{2-13} 
			\textbf{Models (backpropagation, backpropagation-free, ours, ours)} & \textbf{\textit{l}}& \textbf{\textit{nl}}& \textbf{\textit{fs (50)}} & \textbf{\textit{l}}& \textbf{\textit{nl}}& \textbf{\textit{fs (50)}}&
			\textbf{\textit{l}}& \textbf{\textit{nl}}& \textbf{\textit{fs (50)}}& \textbf{\textit{l}}& \textbf{\textit{nl}}& \textbf{\textit{fs (50)}}\\
			\hline
			VAE \cite{hjelm2018learning}	&$54.45$ (SVM)    &$60.71$     &-     &-   &$37.21$     &-     &-    &$18.63$     &-   &-    &-     &-   \\
			BiGAN \cite{hjelm2018learning}	&$57.52$ (SVM)    &$62.74$     &-     &-    &$37.59$     &-     &-    &$24.38$     &-   &-    &-     &-   \\
			DIM(L) (maximum reported accuracy in the paper) \cite{hjelm2018learning}	&$64.11$ (SVM)    &$80.95$     &-     &-    &$\mathbf{49.74}$     &-     &-    &$\mathbf{38.09}$     &-   &-    &-     &-   \\
			AET-project (best single model accuracy we know) \cite{DBLP:journals/corr/abs-1901-04596}	&$\mathbf{83.35}$    &$\mathbf{90.59}$      &-     &-    &-     &-     &-    &-     &-   &-    &-     &-   \\	
			\hline
			SOMNet \cite{10.1007/978-3-030-03493-1_87}	&$78.57$ (SVM)   &-     &-     &-    &-     &-     &-    &-     &-  &-    &-     &- \\
			K-means (Triangle) (best single model accuracy we know) \cite{pmlr-v15-coates11a}	&$\mathbf{79.60}$ (SVM)   &-     &-     &-    &-     &-     &-    &-     &-  &-    &-     &- \\
			\hline
			D-CSNN-B1	&$73.73$    &$77.83$     &$\mathbf{41.59}$       &$45.17$    &$49.80$     &$27.29$   &$13.52$    &$3.56$     &$13.70$ 	&$68.92$    &$70.69$     &$\mathbf{49.99}$\\
			D-CSNN-B1-Aug &$\mathbf{75.68}$    &$\mathbf{81.55}$     &-     &$\mathbf{46.82}$      &$\mathbf{52.66}$     &-     &$13.32$    &$\mathbf{5.20}$      &-  	&$\mathbf{70.80}$    &$\mathbf{75.29}$    &-\\
			\hline
			D-CSNN-B1-Cifar10	&$\mathbf{73.73}$    &$77.83$     &$\mathbf{41.59}$     &$45.68$    &$49.70$     &$\mathbf{27.96}$     &$\mathbf{14.36}$    &$2.03$     &$13.46$    &$65.34$    &$65.59$  &$47.65$	\\
			D-CSNN-B1-Cifar100	&$71.60$    &$76.46$     &$41.43$     &$45.17$    &$49.80$     &$27.29$     &$13.57$    &$1.61$     &$\mathbf{13.72}$  &$63.79$    &$65.13$     &$45.68$  \\
			\hline
		\end{tabular}
		\label{performances}
	\end{center}
\end{table*}
	
	\subsection{Visual Interpretations}
\begin{figure}[tbp]
		\centerline{\includegraphics[width=\columnwidth]{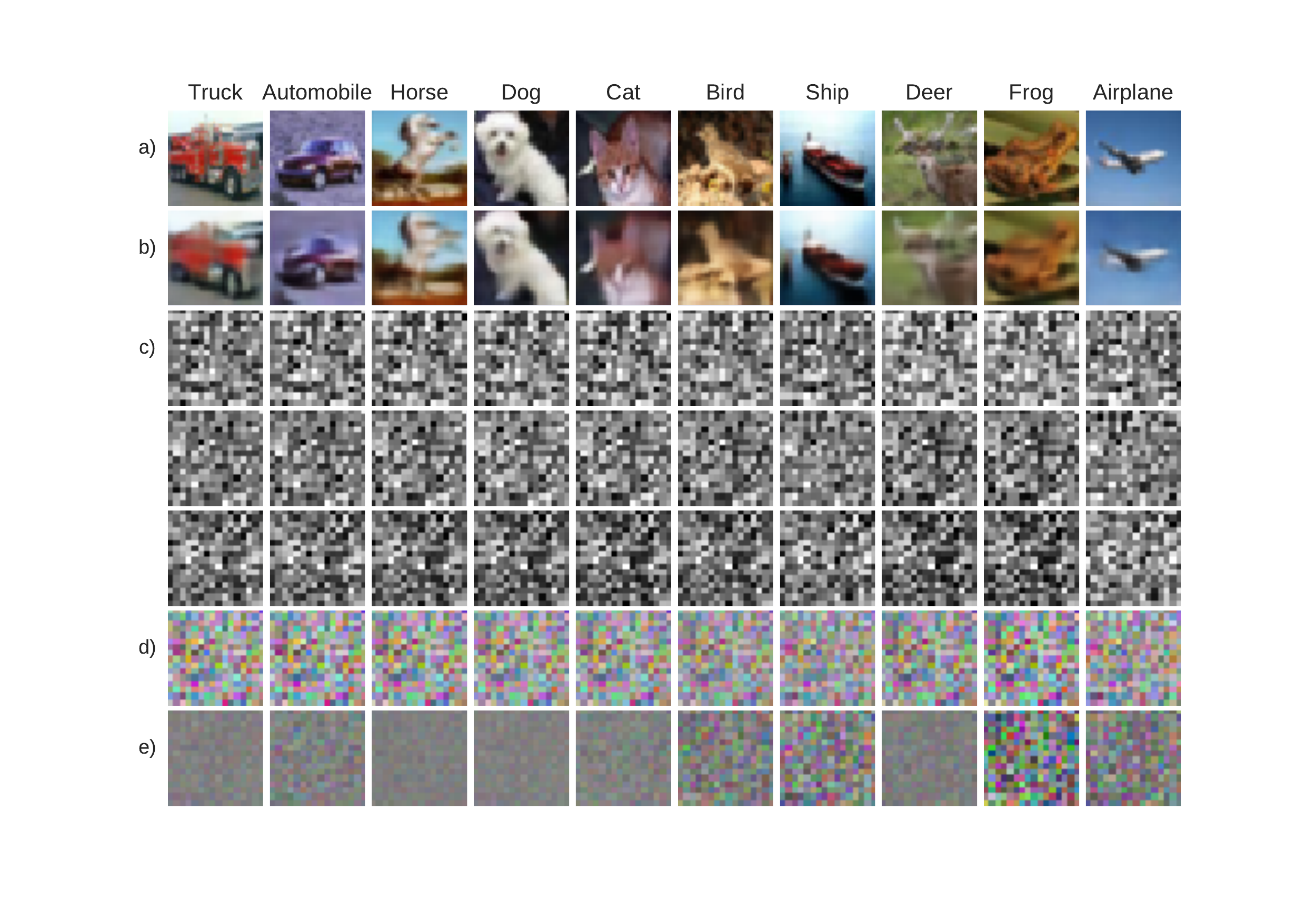}}
		\caption{a) The Cifar10 input image. b) The reconstruction using the representation of the D-CSNN-B1 (no augmentation). c) The average representation per map for each class, formed by taking the mean over the spatial dimension of each test sample's representation and reshaping the resulting vector to SOM-grid shape. d) RGB image of the average representation (three maps lead to an RGB image). e) Differences of several average representations achieved through element-wise subtracting the classes average representation from the classes average representation on the right hand side. Best viewed in color.}
		\label{representations}
	\end{figure}

	Fig. \ref{representations} row b) shows that the input images can be reconstructed from the representation showing a high degree of detail.
	Furthermore, one can see in rows c) and d) that ``similar'' classes lead to similar average representations, especially there exist similar spots with high average neuron activities. The classes are sorted according to their $L_1$ distance from the truck class and we can see that the car class is most similar, followed by the horse class. Row e) shows remaining distinguishable differences between average representations of the class and the neighboring class on the right hand side (e.g., the first entry equals the truck minus car average representation). 
	
	\section{Discussion and Conclusion}
	In this work we introduced the modular building blocks of CSNNs to learn representations in an unsupervised manner without Backpropagation. With the proposed CSNN modules — which combine CNNs, SOMs and Hebbian learning of masks — a new, alternative way for learning unsupervised feature hierarchies has been explored. Along the way we discussed the Scalar Bow Tie Problem and the Objective Function Mismatch: two problems in the field of Deep Learning, which we belief can be potentially solved together and are an interesting direction for future research. 
	\bibliographystyle{IEEEtran}
	\bibliography{IEEEabrv,main}

\appendix
\label{appendixvis}
\begin{figure*}[htbp]
	\centerline{\includegraphics[width=\textwidth]{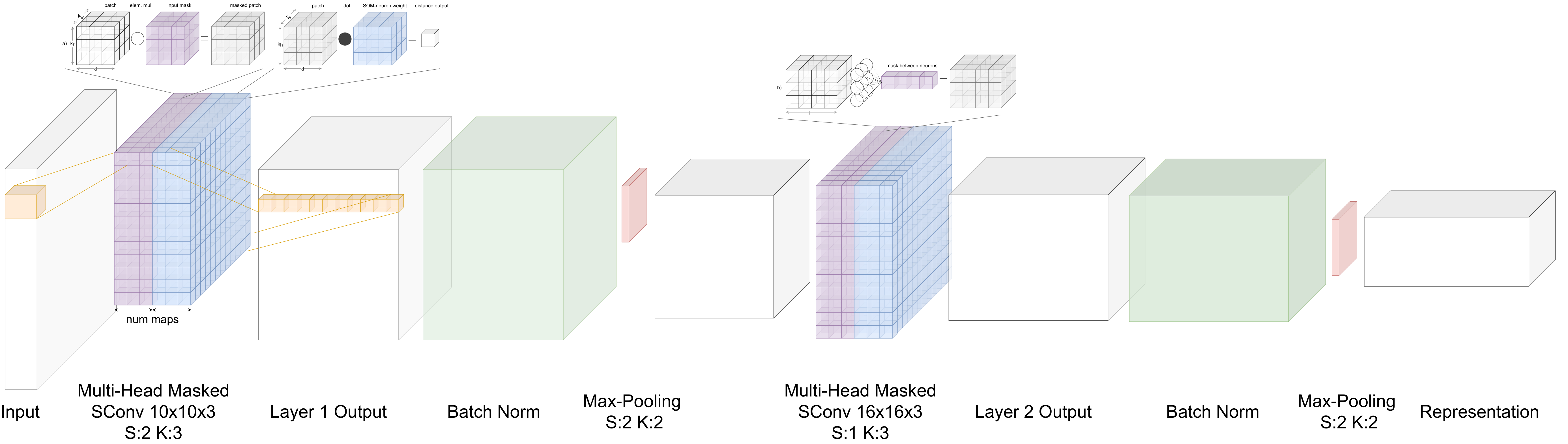}}
	\caption{Detailed architecture of the S-CSNN with three SOM-maps. The model consists of two CSNN-layers, each is followed by a batch norm and a max-pooling layer to halve the spatial dimension of the output tensor. The first layer uses a $10\times10\times3$ SOM-grid (3 heads with 100 SOM-neurons), stride $2\times2$ and input mask weights. The second layer uses a $16\times16\times3$ SOM-grid, stride $1\times1$  and mask weights between neurons. Both layers use a kernel size of $3\times3$ and padding type ``same''. Best viewed in color.}
	\label{csnnarchitecture}
\end{figure*}
\begin{figure*}[htbp]
	\centerline{\includegraphics[width=\textwidth]{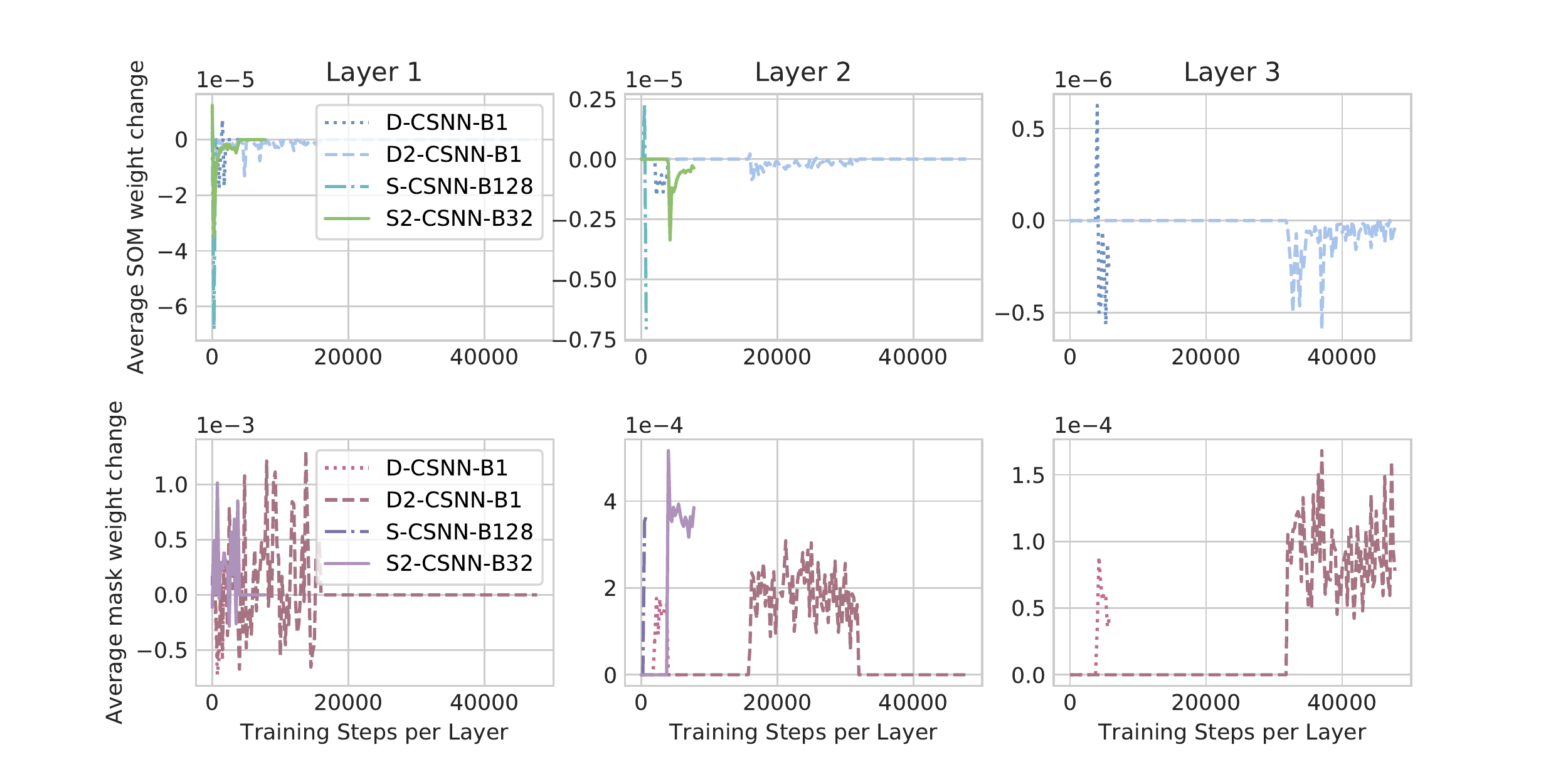}}
	\caption{The first row shows the average SOM-weight changes during training of each layer. We can see that each layer was trained bottom up in its defined training interval. The SOM-weights show convergence, which takes longer in deeper layers (and possibly could be improved when learning rates and neighborhood coefficients change over the course of the training). We argue that the reasons for the slower convergence of deeper layers are the increased kernel and SOM-map sizes. The second row shows average mask weight changes. Best viewed in color.}
	\label{weightchanges}
\end{figure*}
\begin{figure*}[htbp]
	\centerline{\includegraphics[width=\textwidth]{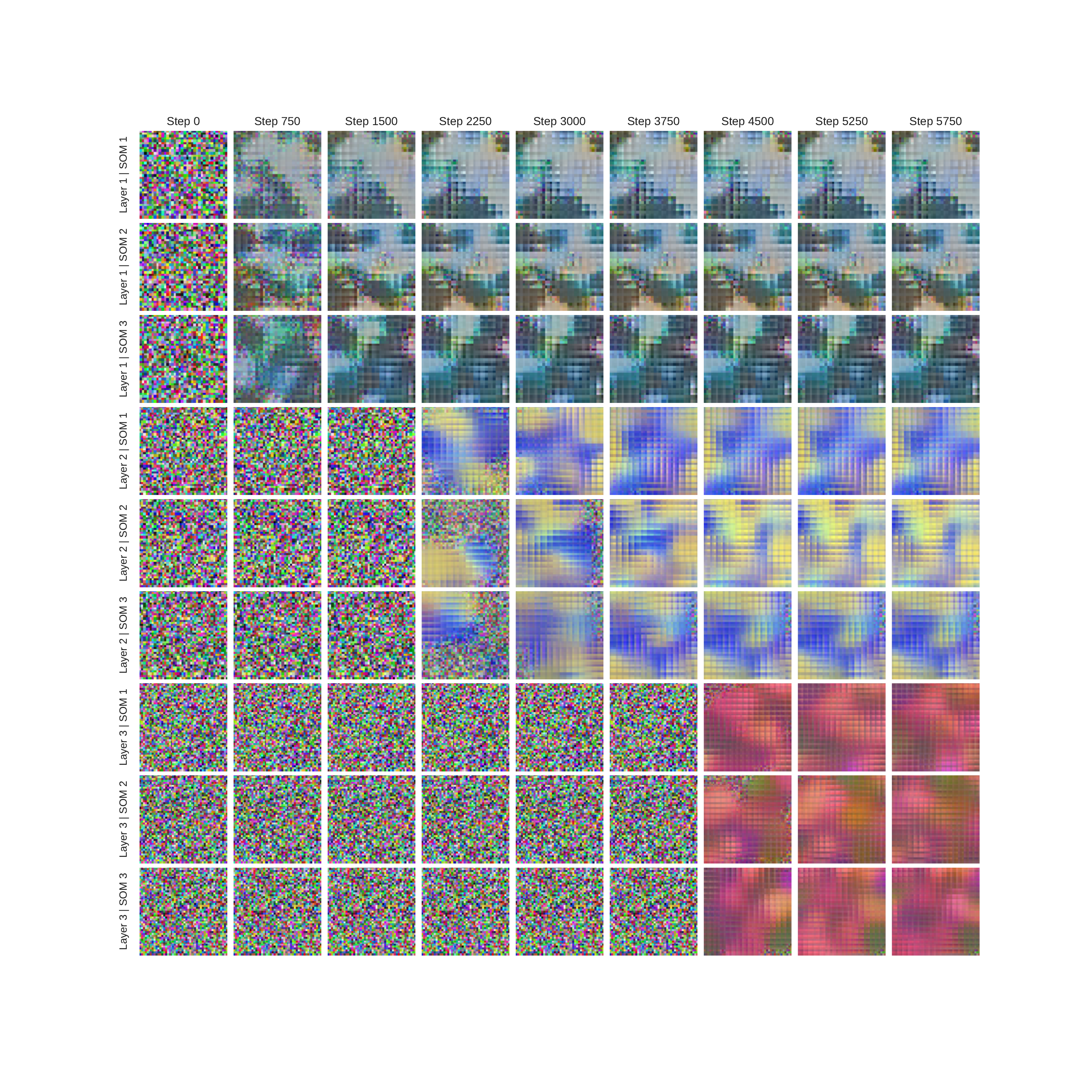}}
	\caption{The layers SOM-weights shown at their SOM-grid and map position throughout the training process of the D-CSNN. Each SOM-image is formed by taking the weights from all SOM-neurons of the map at the current training step, reshaping them to the patch shape (3D) and forming an image by placing them according to the corresponding SOM-neuron coordinates in the SOM-grid. For deeper layers we only show a slice of depth three through the SOM-weights, because it is hard to visualize kernel (SOM-weight) depths larger then three (e.g., the depth 300 when there are three $10\times10$ SOMs in the previous layer). We can see that all SOM-weights were trained bottom up in their defined training intervals. Furthermore, this clearly figure shows some kind of self-organization. We want to note that the individual SOM-weights in layer 1 do not necessary represent color, as shown in the following Fig. \ref{bmuimages}, \ref{bmuimages2} and \ref{bmuimages3}; the reason for that are the mask weights. Best viewed in color.}
	\label{somweights}
\end{figure*}
\begin{figure*}[htbp]
	\centerline{\includegraphics[width=\textwidth]{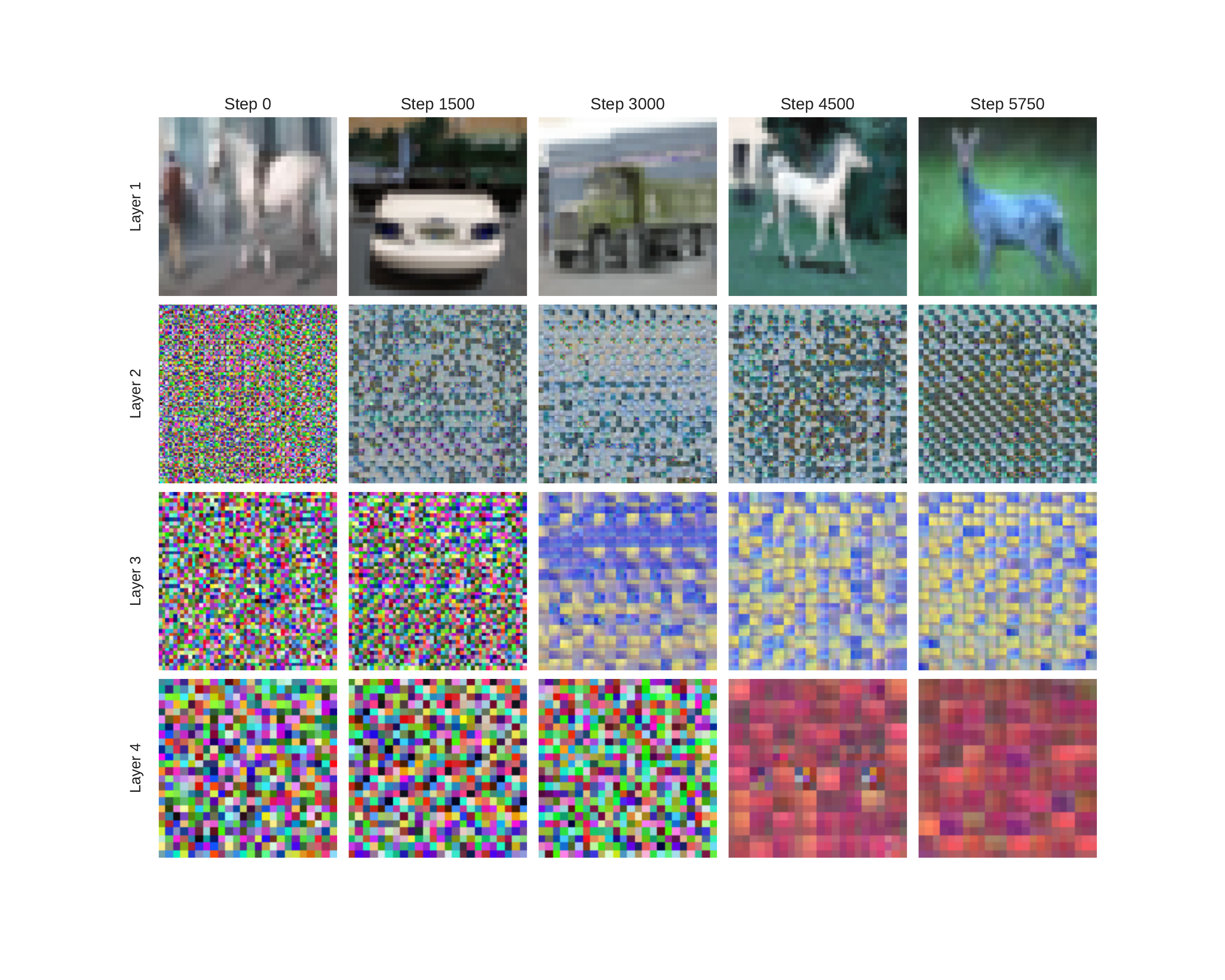}}
	\caption{BMU-images for each layer of the D-CSNN created by replacing each patch with the SOM-weight of the BMU, which is reshaped into the 3D patch shape. For deeper layers we only show a slice of depth three through the BMU-weight, because its hard to visualize kernel depths larger then three. Since the D-CSNN-B1 contains three maps, we use the best BMU from all maps of the layer, but it is also possible to inspect individual maps. We can see that the BMUs for deeper layers where noise for the first images, because the layers had not been trained yet. After each layer was trained, patterns emerged. For layer 1 one can see patterns corresponding to the input regions content. These patterns become less interpretable in deeper layers. Image 5 of layer 2 shows that the SOM-weights are not only sensitive to colors, because different backgrounds result in similar BMU patterns. Best viewed in color.}
	\label{bmuimages}
\end{figure*}
\begin{figure*}[htbp]
	\centerline{\includegraphics[width=\textwidth]{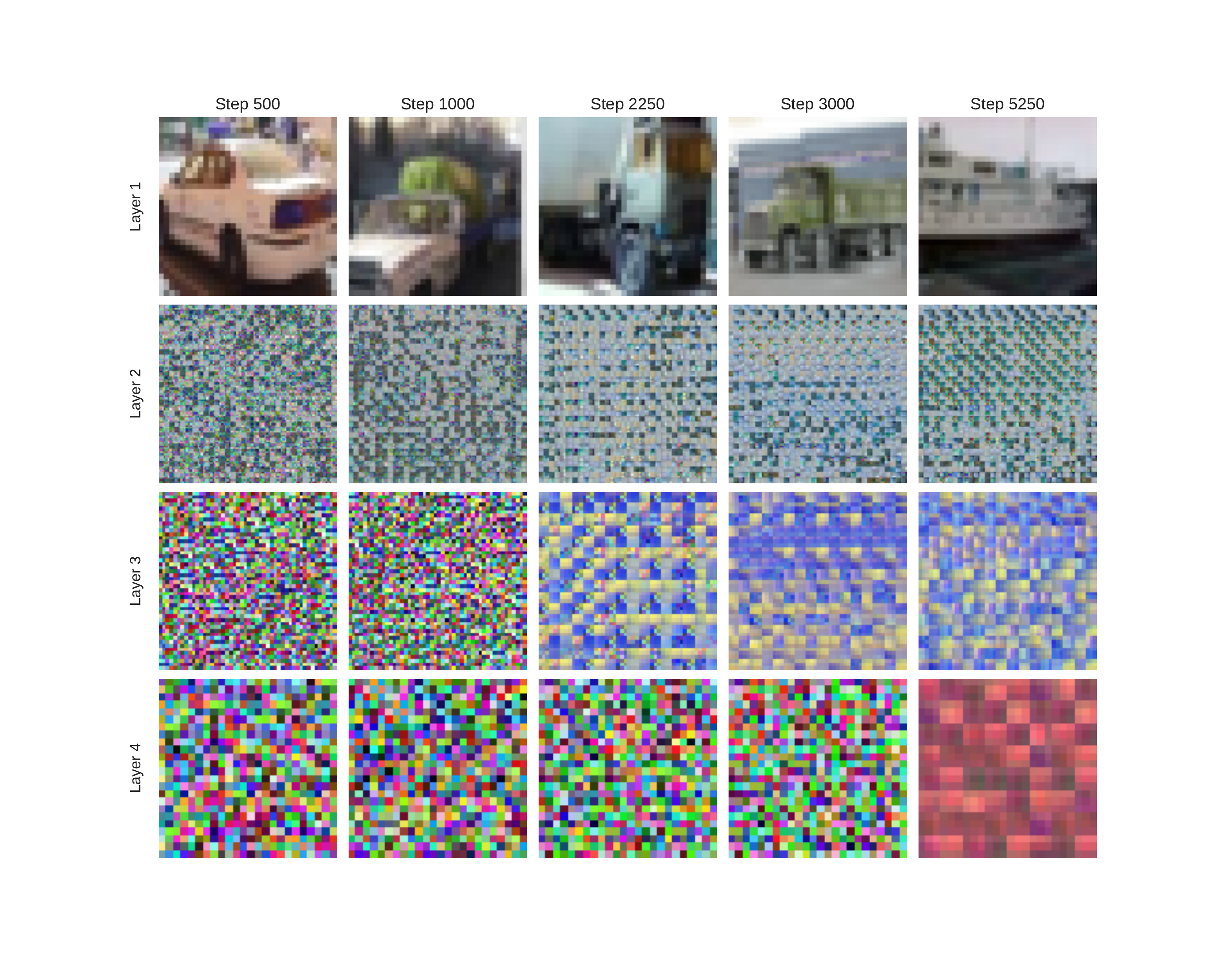}}
	\caption{More BMU-images. Best viewed in color.}
	\label{bmuimages2}
\end{figure*}
\begin{figure*}[htbp]
	\centerline{\includegraphics[width=\textwidth]{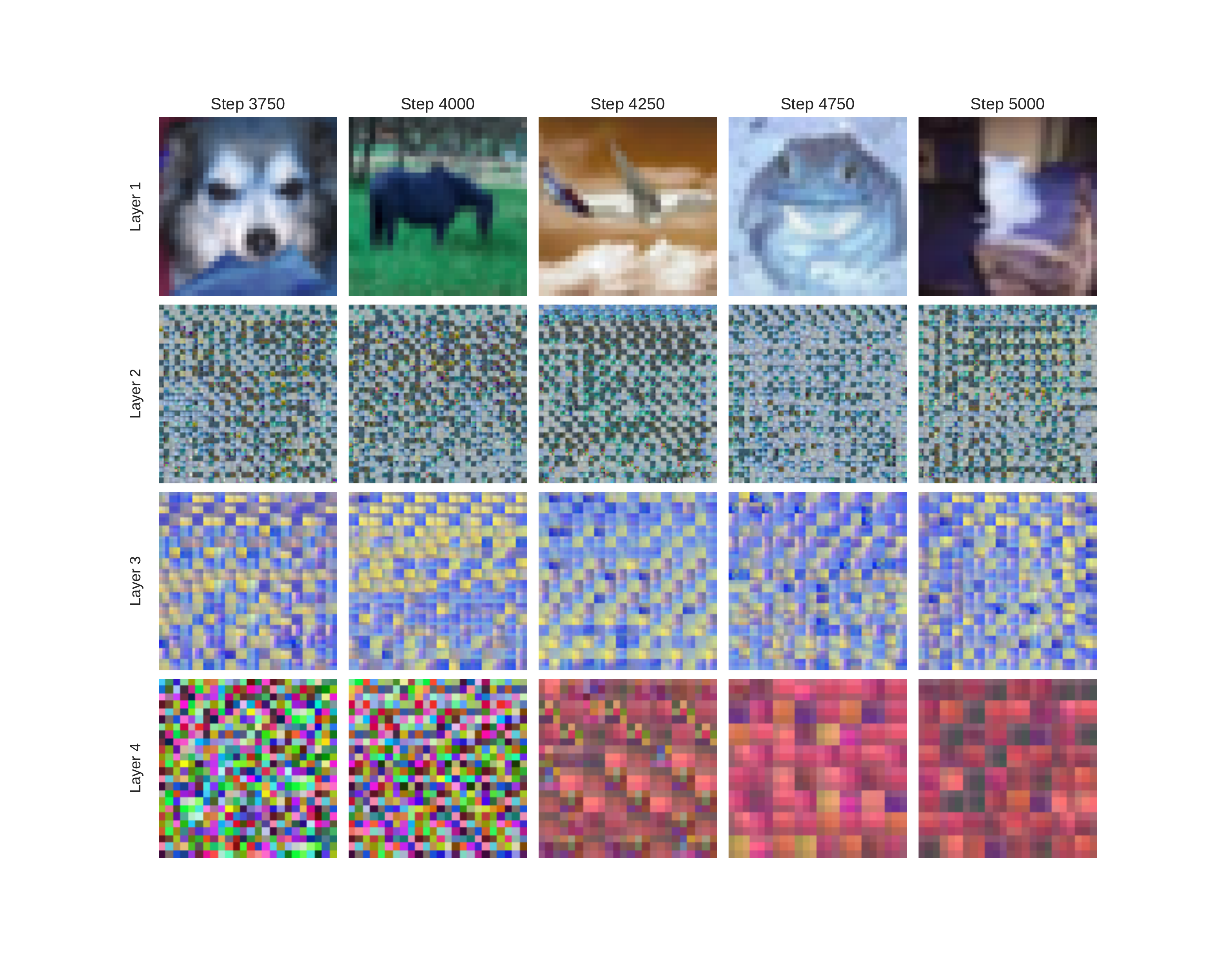}}
	\caption{More BMU-images. Best viewed in color.}
	\label{bmuimages3}
\end{figure*}

\begin{figure*}[htbp]
	\centerline{\includegraphics[width=\textwidth]{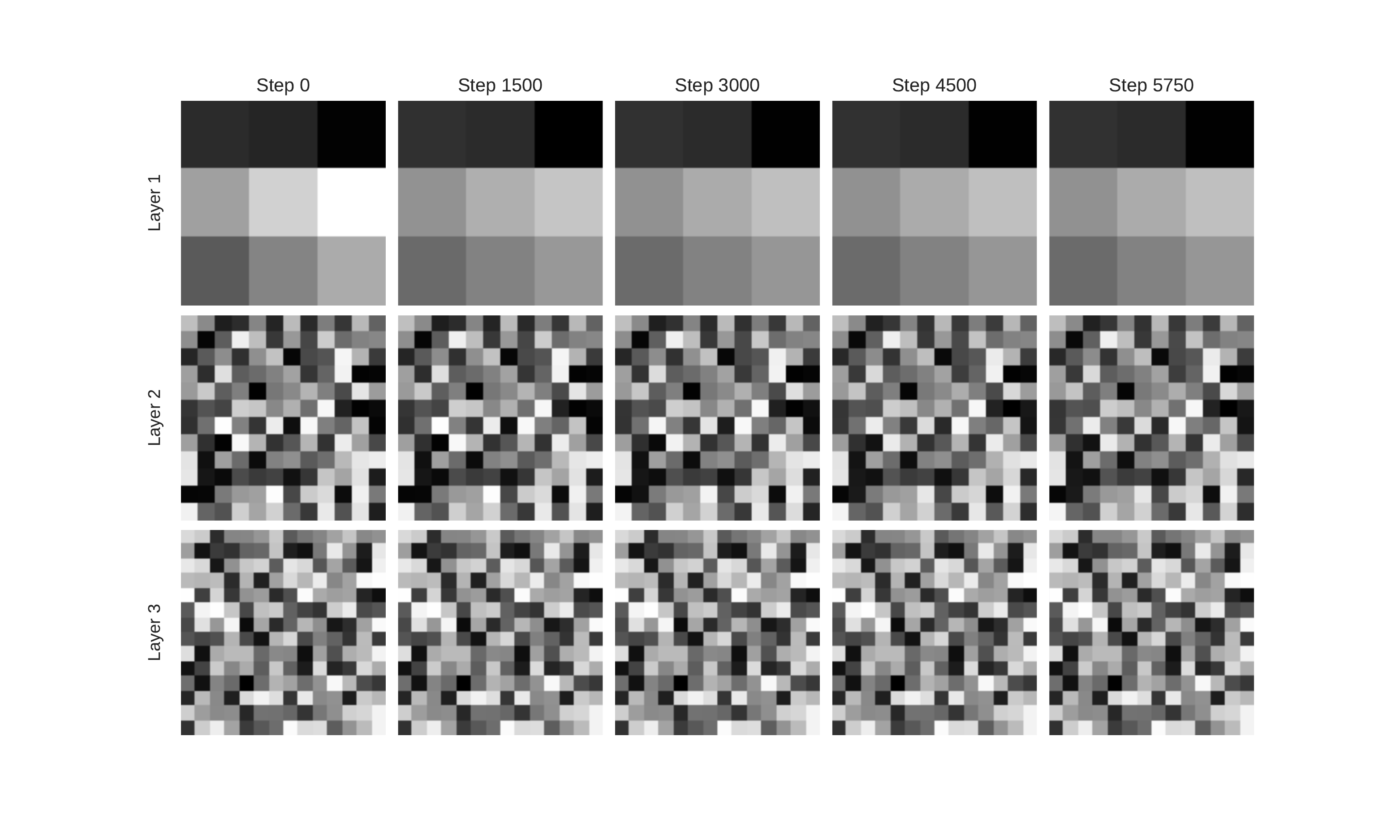}}
	\caption{Shows a slice of depth one through the mask values of one mask for each layer, where white corresponds to $1.0$ and black to $-1.0$. One can see that throughout the training process these values change slightly, indicating that the local mask learning chooses preferred inputs.}
	\label{masks}
\end{figure*}
\begin{figure*}[htbp]
	\centerline{\includegraphics[width=\textwidth]{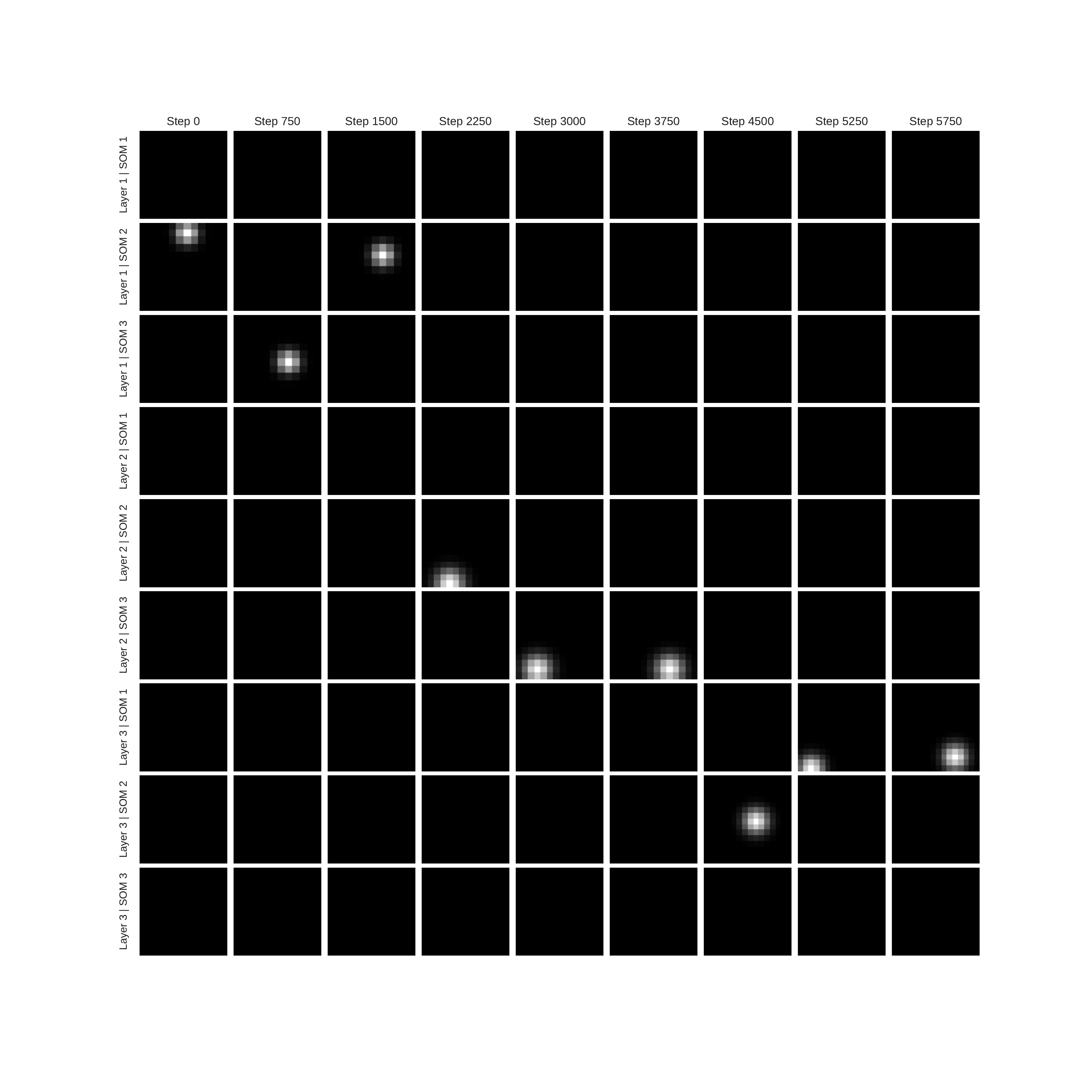}}
	\caption{Shows the neighborhood coefficients of one spatial activation for each layer and map, where white corresponds to $1.0$ and black to $0.0$. To create one image the flat vector containing the neighborhood coefficients is reshaped to the 2D SOM-grid shape. One can see that the neighborhood is Gaussian and that only the map with the best BMU was updated. Furthermore, one can see that layers are only trained in their defined training intervals $[y, x]$}
	\label{mapsCoeffs}
\end{figure*}
\begin{figure*}[htbp]
	\centerline{\includegraphics[width=0.8\textwidth]{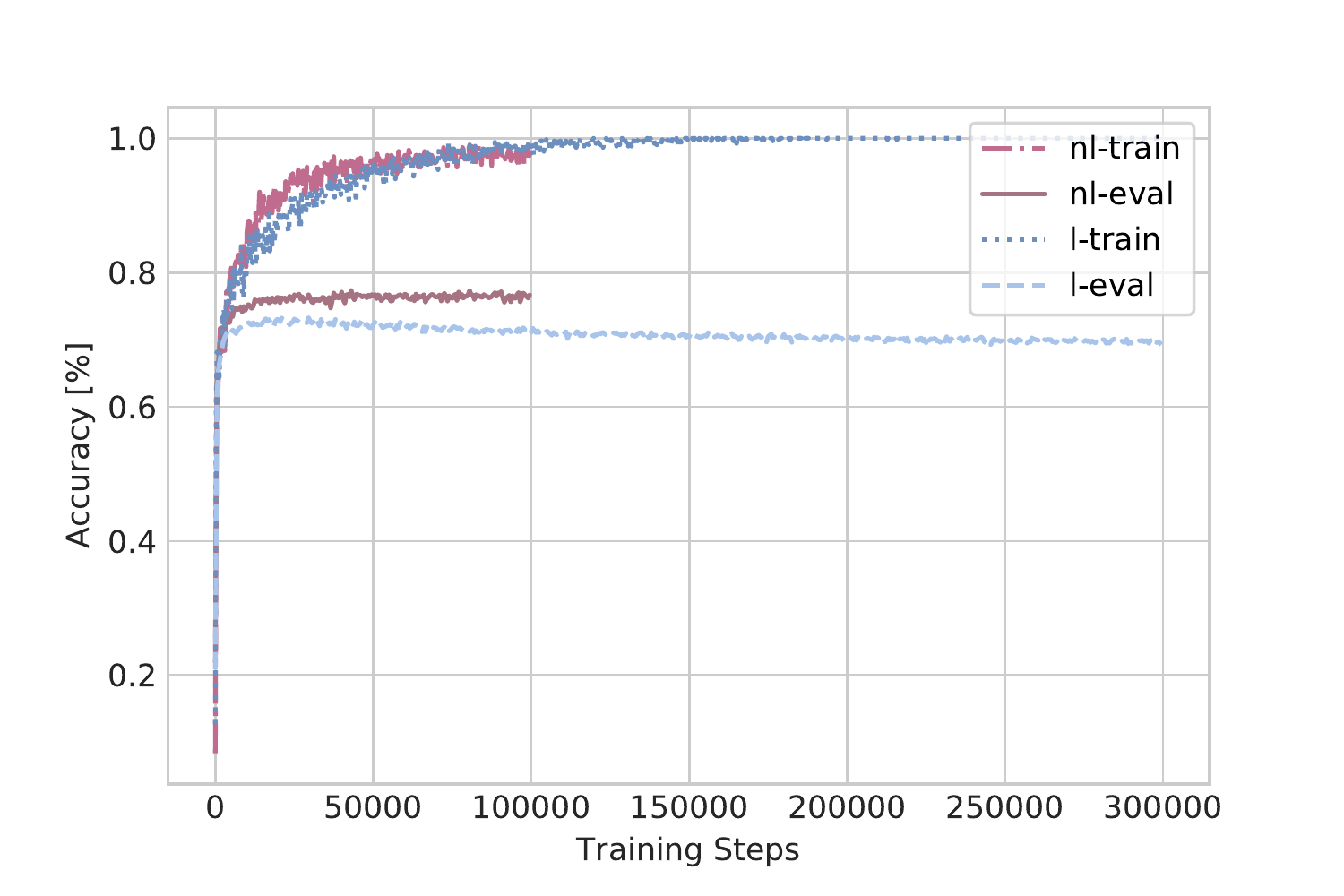}}
	\caption{The accuracy curves for the linear and nonlinear classifier training on the D-CSNN-B1 representations.}
	\label{learnincurves}
\end{figure*}
\begin{figure*}[htbp]
	\centerline{\includegraphics[width=0.8\textwidth]{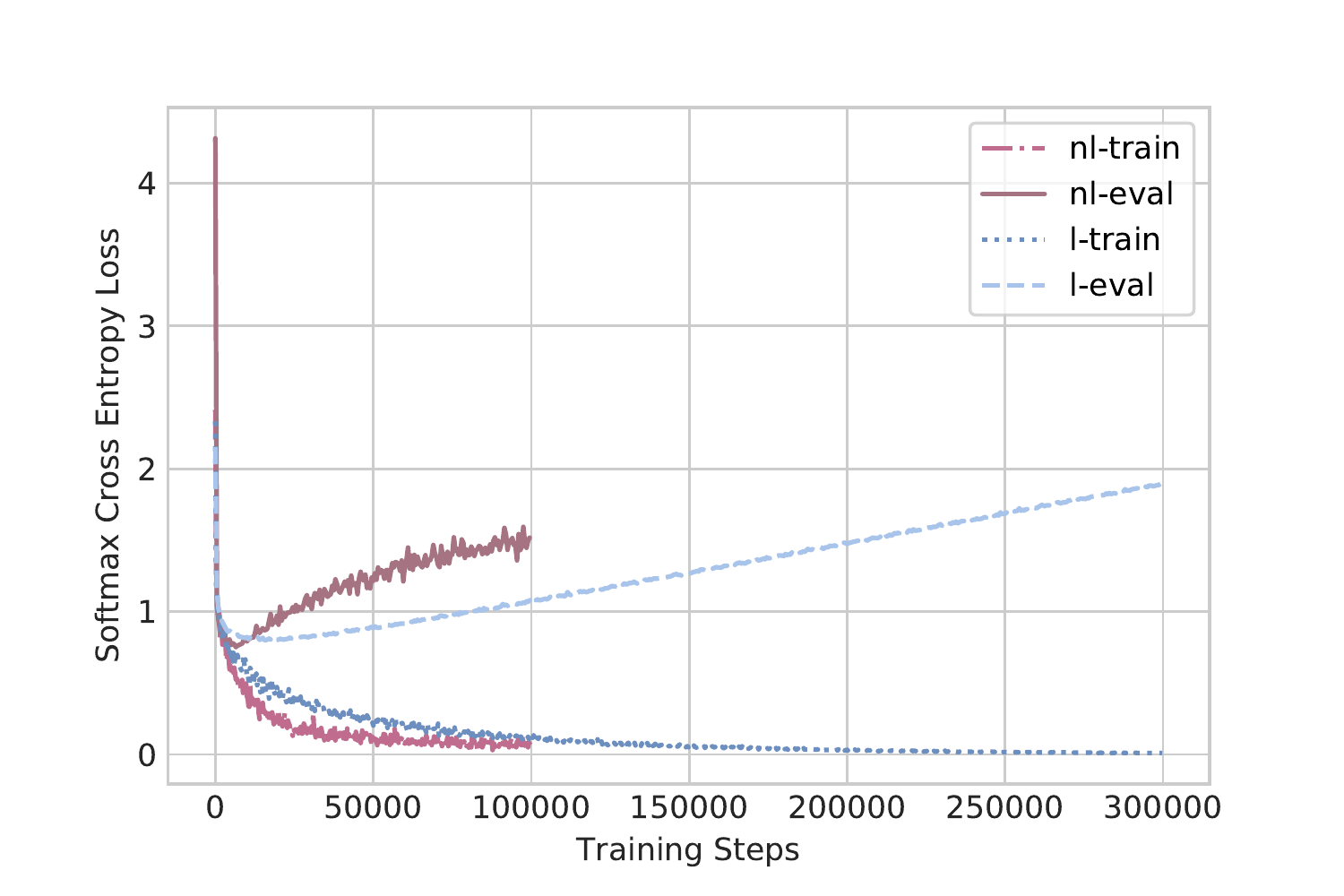}}
	\caption{The loss curves for the linear and nonlinear classifier training on the D-CSNN-B1 representations.}
	\label{losscurves}
\end{figure*}
\begin{figure*}[htbp]
	\centerline{\includegraphics[width=\textwidth]{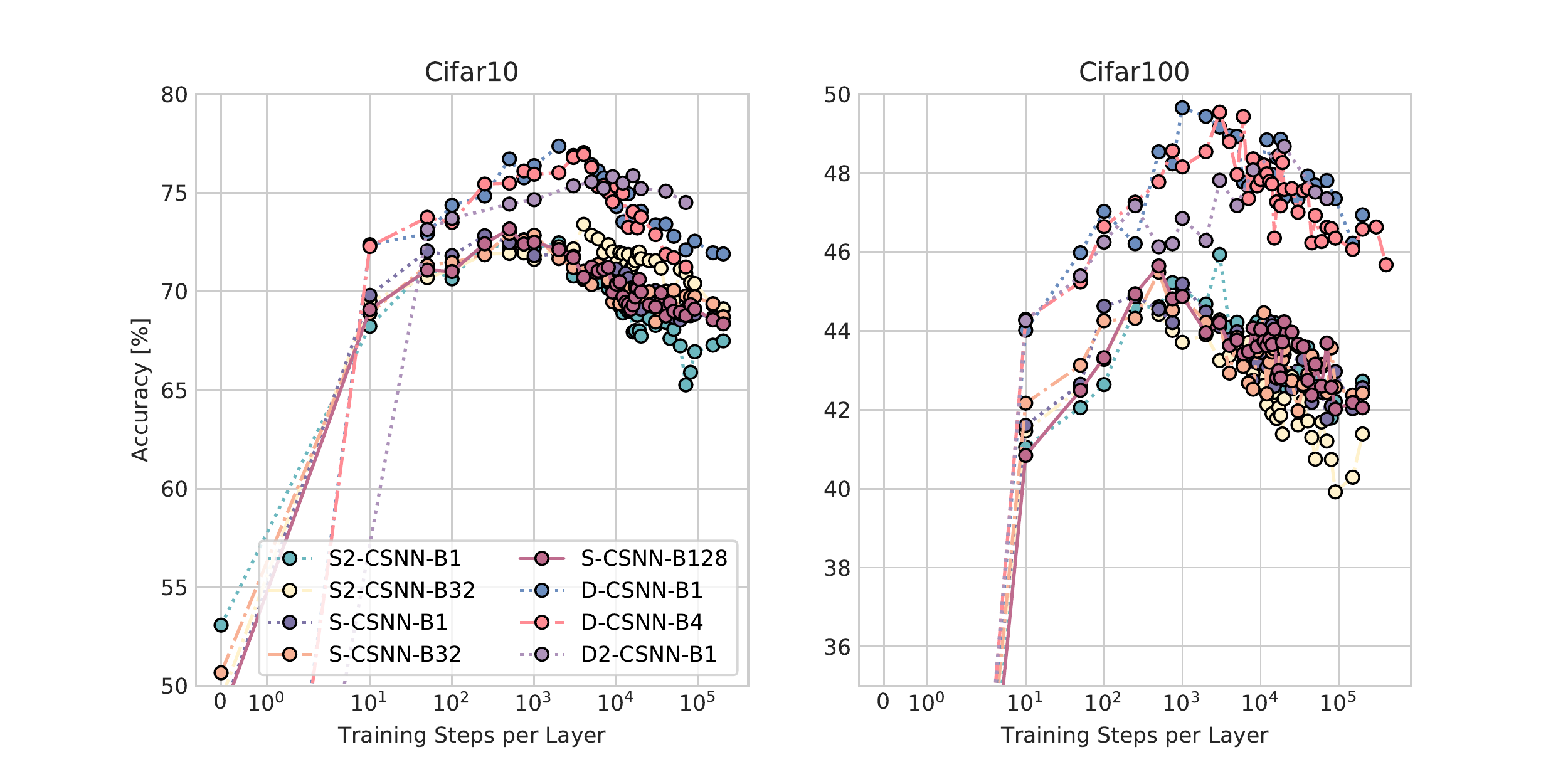}}
	\caption{Accuracies of our models trained with our local mask learning rules (S/D for \eqref{eqn:similar_to_gha} and S2/D2 for \eqref{eqn:similar_to_gha_with_restricted_summation}) and different batch sizes B. Every point corresponds to an nl classifier trained on the representation of the CSNN, the layers of which where trained layer-wise for the steps shown on the x-axes. Best viewed in color.}
	\label{ofmbig}
\end{figure*}
\begin{figure*}[htbp]
	\centerline{\includegraphics[width=\textwidth]{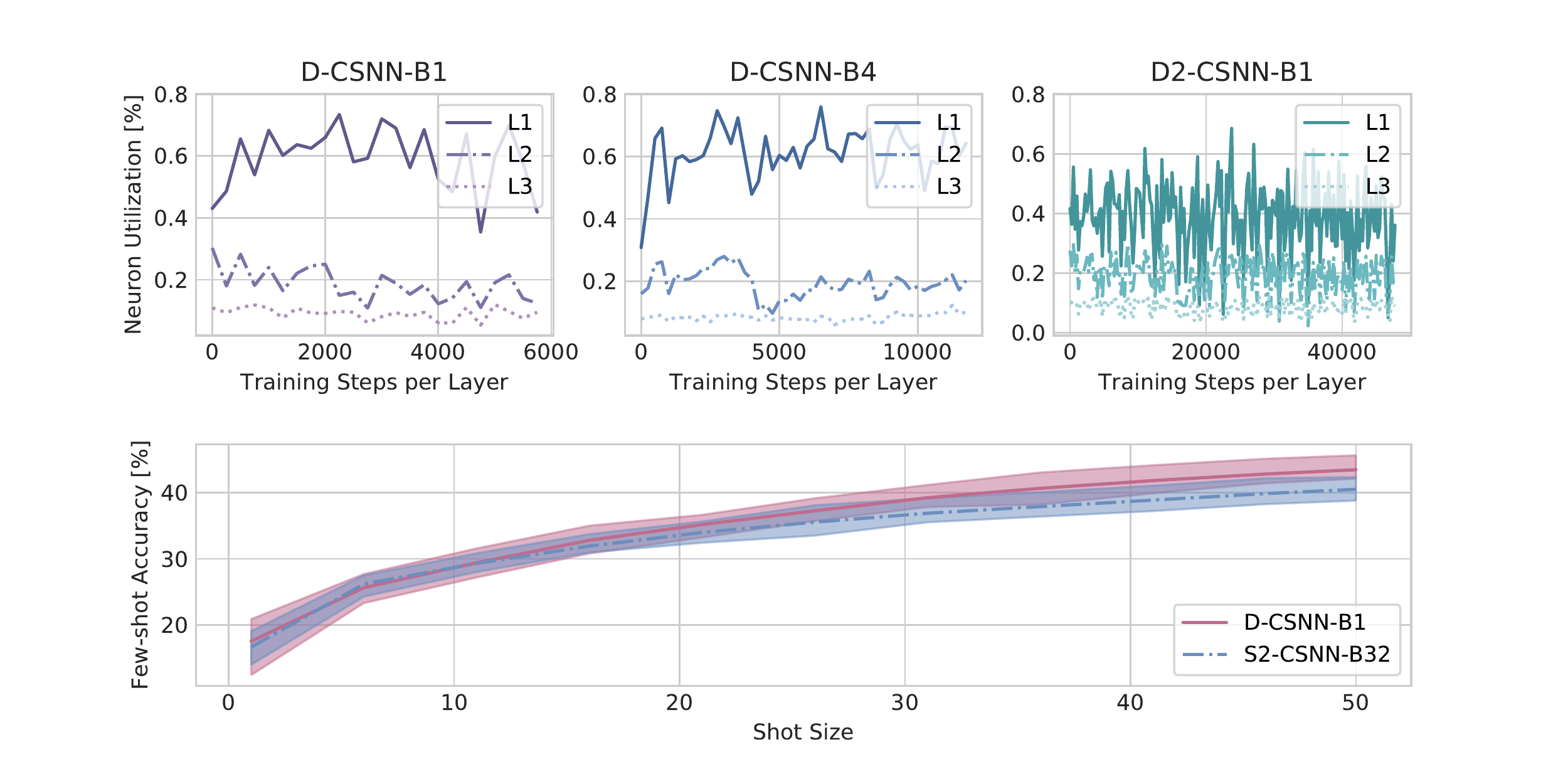}}
	\caption{(top) The neuron utilization of models trained with various batch sizes and mask learning rules on Cifar10. (bottom) The 10fold few-shot classifier accuracy of these models with increasing shot sizes from $1$ to $50$ on Cifar10.}
	\label{utilAndFsbig}
\end{figure*}
\end{document}